\documentclass{article}

\usepackage{arxiv}

\usepackage[utf8]{inputenc} 
\usepackage[T1]{fontenc}    
\usepackage{hyperref}       
\usepackage{url}            
\usepackage{booktabs}       
\usepackage{amsfonts}       
\usepackage{nicefrac}       
\usepackage{microtype}      
\usepackage{lipsum}
\usepackage{fancyhdr}       
\usepackage{graphicx}       
\graphicspath{{media/}}     

\usepackage{amsmath}
\usepackage{mathtools}

\usepackage{biblatex} 
\usepackage{blindtext}
\usepackage{enumitem}
\usepackage[dvipsnames]{xcolor}
 
\usepackage{tikz}
\usetikzlibrary{positioning,shapes,snakes}

\title{
The Neuro-Symbolic Brain
}
\author{Robert Liz\'ee \\ \href{mailto:robert.lizee@gmail.com}{robert.lizee@gmail.com}}
\date{\today}

\begin{document}

\maketitle

\begin{abstract}
    Neural networks promote a distributed representation with no clear place for symbols. 
    Despite this, we propose that symbols are manufactured simply by training a sparse random noise as a self-sustaining attractor in a feedback spiking neural network. 
    This way, we can generate many of what we shall call prime attractors, and the networks that support them are like registers holding a symbolic value, and we call them registers. 
    Like symbols, prime attractors are atomic and devoid of any internal structure. Moreover, the winner-take-all mechanism naturally implemented by spiking neurons enables registers to recover a prime attractor within a noisy signal. 
    Using this faculty, when considering two connected registers, an input one and an output one, it is possible to bind in one shot using a Hebbian rule the attractor active on the output to the attractor active on the input. 
    Thus, whenever an attractor is active on the input, it induces its bound attractor on the output; even though the signal gets blurrier with more bindings, the winner-take-all filtering faculty can recover the bound prime attractor. 
    However, the capacity is still limited. It is also possible to unbind in one shot, restoring the capacity taken by that binding. 
    This mechanism serves as a basis for working memory, turning prime attractors into variables. 
    Also, we use a random second-order network to amalgamate the prime attractors held by two registers to bind the prime attractor held by a third register to them in one shot, de facto implementing a hash table. 
    Furthermore, we introduce the register switch box composed of registers to move the content of one register to another. 
    Then, we use spiking neurons to build a toy symbolic computer based on the above.
    The technics used suggest ways to design extrapolating, reusable, sample-efficient deep learning networks at the cost of structural priors. 
\end{abstract}

\section{Introduction}
    Deep learning, a technology inspired by our understanding of the brain, has recently had considerable successes in artificial intelligence. 
    However, these successes have led to criticisms on its perceived shortcomings in its ability to address full human cognition (Gary Marcus \cite{marcus2018deep}). 
    Jacob L Russin and al. in \cite{dlneedspfcortex} argue that deep learning needs a frontal cortex. 
    Yosha Bengio in \cite{aaai2020keynotes} acknowledges some challenges with deep learning.
    He articulates it through the dichotomy of System 1 and System 2 thought systems as described by Daniel Kahneman's book \emph{Thinking, Fast and Slow} \cite{Kahneman11}.
    He argues that deep learning is adequate for System 1 tasks but maintains that while System 2 tasks are in the realm of classical artificial intelligence's elusive goal, deep learning could evolve to handle them.
    Gary Marcus advocates in \cite{marcus2020decade} for a hybrid system combining deep learning technics with symbolic manipulation as an avenue to mitigate deep learning limitations,
    noting that successful deep learning systems like alphaGo are hybrid systems. 
    Marcus further proposes that the ability to do symbolic operations over variables is essential but missing from current deep learning.
    This work is a response to this missing link. 
    Here, we propose that feedback neural networks of spiking neurons can indeed do symbolic operations over variables in 
    a limited but highly relevant manner. 

    Neural networks promote a distributed representation (see
    Geoffrey E Hinton and al. in Hinton \cite{HintonMcClellandRumelhart86} and \cite{hinton1986learning})
    which seems incompatible with the atomic nature of symbols.
    We propose that symbols are created by training a sparse random noise pattern as an attractor. 
    Given that the attractor's internal structure is noise, they possess no intrinsic information like symbols, and they cannot be decomposed into substructures, making them, like symbols, atomic. 
    We call these attractors prime attractors.
 
    David S. Touretzky and al. in \cite{touretzky:symbols} proposed to use a somewhat random representation for triplets of symbols with a rich internal structure tailored to solve the problem set forth by their work.
    Here, we propose a per symbol structureless random representation usable in a generic fashion.

    Our objective is to explore how the human brain could do symbolic processing using spiking neural networks to represent the brain.
    Neural networks are known to be Turing complete, as shown in \cite{siegelman:computational}.
    However, regarding our objective, the Turing completeness proof relies on encoding an unbounded binary stack as a rational number into a neuron.
    This encoding is unrealistic for the brain.

    Pursuing our objective, we uncovered the means to do a one-shot symbolic memory with prime attractors.
    In order to control the memory, we add to our spiking neural network two types of connections between neurons: one that triggers a binding behavior in the output neuron for memorization and one that triggers an unbinding behavior for forgetting. 
    We built a toy symbolic computer around that memory with a spiking neural network simulated on a conventional computer.
    To do so, we had to introduce another another crucial component, the register switch box and argue for its physiological plausibility.
    Our symbolic computer differs from a conventional one because it only manipulates symbols; it does not have low-level support for integers or floating-point numbers.
    The registers of the symbolic computer can only hold symbols.
    One can represents digits with symbols, but for integers, our symbolic computer uses linked list of digits to represent them.
    Our symbolic computer is capable of executing programs written in an assembly language.

    Our symbolic computer also differs from the neural Turing machine \cite{graves2014neural} and the differentiable neural computer \cite{graves2016hybrid} which are forms of memory augmented neural networks.
    Our one-shot symbolic memory implementation is with the spiking neurons themselves, and it is not a module added to a neural network.
    Our memory is restricted to symbols.
    Our architecture is not end-to-end differentiable but instead is reminiscent of a conventional computer that needs to be programmed.
    We show how we did this and then argue how our symbolic computer can evolve and its implication in helping understand how the brain reaches full human cognition beyond today's deep learning networks, revealing the relevance of research into hybrid systems.
    We also argue that some of the technics develop to build our symbolic computer may be used in designing deep learning neural networks to augment their reusability, their ability to extrapolate, and their sample efficiency at the cost of architectural priors.

    \begin{itemize}
        \item Section \ref{SymbolicComponents} introduces some symbolic components required to build our symbolic computer. 
        \begin{itemize}
            \item To understand the brain as a symbolic computer, Section \ref{FeedbackSpikingNeuralNetwork} presents an \textbf{assemblage of interconnected feedback spiking neural networks} as our framework to implement our symbolic computer.
            \item Section \ref{DistributedRepresentation} discusses how \textbf{symbols} are represented in a \textbf{distributed representations}.
            \item Section \ref{PrimeAttractor} presents \textbf{prime attractors} and \textbf{registers} that serve as the key concepts for this work.
            A prime attractor have a symbol as fix point.
            It serves to filter out the noise around a symbol.
            A \emph{register} is a network that holds a prime attractor.
            Registers are the fundamental building block with which we build the other components used to make a symbolic computer.
            \item Section \ref{OneShotSymbolicMemory} shows our \textbf{one-shot symbolic memory} which can bind/unbind in one shot a symbol in one register to another symbol in another register through a network connecting the two registers.
            This one-shot symbolic memory enables a prime attractor to act as a variable and is the foundation of this work.
            \item Section \ref{Hashtable} extends the one-shot symbolic memory network to build a \textbf{hash table}.
            \item Section \ref{RelatingToTheBrain} discusses the \textbf{relation to the human brain} in terms of how many prime attractors and bindings a system can support.
        \end{itemize}

        \item Section \ref{SymbolicFramework} extends and discusses the framework introduced in section \ref{FeedbackSpikingNeuralNetwork}.
        \begin{itemize}
            \item Section \ref{SymbolicFrameworkUpgrade} \textbf{upgrades} our symbolic framework, notably with the use of control connections.
            \item Section \ref{Encodings} discusses about the type of \textbf{encodings} used in the framework.
            \item Section \ref{Connections} elaborates on the type of \textbf{connections} present in our symbolic computer.
            \item Section \ref{RegisterSwitchbox} then presents a \textbf{register switch box} composed of several registers that can move the value of a register into another register and argue for its physiological plausibility.
        \end{itemize}

        \item With these pieces, section \ref{buildingacomputer} proceeds to \textbf{build a symbolic computer} with a full-blown assembly language.
        \begin{itemize}
            \item Section \ref{AssemblyLanguage} presents the \textbf{assembly language} used.
            \item Section \ref{Architecture} presents the \textbf{architecture} of the symbolic computer.
            \item Section \ref{CommonSequences} shows \textbf{common sequences of instructions}, which give a sense on how to program this symbolic computer.
            \item Section \ref{ComputerInitialization} discusses the various steps to \textbf{initalize} our symbolic computer.
            \item Section \ref{ComputerTests} looks at the \textbf{results} of the test programs of echoing the input, adding two numbers represented sequence of digits in the input, and counting the number of each digit present in the input.
        \end{itemize}
    
        \item Finally section \ref{discussion} \textbf{discusses} implications of our symbolic computer.
        \begin{itemize}
            \item Section \ref{ComputationalModelOfTheBrain} argues that our symbolic computer can evolve into a neuro-symbolic computer.
            The suggestion follows that these neuro-symbolic computers could serve as a \textbf{computational model of the brain}.
            The dichotomy between neural and symbolic processing is reminiscent of the dichotomy of System 1 and System 2 of Daniel Kahneman's book.
            \item Section \ref{LimitsOfBackpropagation} dicusses the \textbf{limits of backpropagation}.
            \item Section \ref{DeepLearning2.0} presents how technics used for the architecture of the symbolic computer may be used to design deep learning network with \textbf{better reusability, ability to extrapolate, and sample efficiency} at the cost of architectural priors.
            \item Section \ref{HybridSystems} finally argues that in AI, to pursue hybrid systems is consistent with our proposed computational model of the brain and presents one way to conceive them via an \textbf{oracle machine}.
        \end{itemize}

        \item Appendix \ref{ShorttimeMemoryCapacity} further argues mathematically on the capacity of the one-shot symbolic memory.

        \item Appendix \ref{spikingvsnonspiking} exemplifies mathematically the diffence between spiking neural network and non-spiking neural network in filtering noise.

    \end{itemize}

\section{Symbolic components} \label{SymbolicComponents}

    The one-shot symbolic memory is the main component we introduce in this work that enables constructing a symbolic computer with spiking neurons.
    A feedback spiking neural network serves as the framework to support prime attractors and the registers that hold them, which in turn are used to implemented the one-shot symbolic memory.

\subsection{Feedback spiking neural networks} \label{FeedbackSpikingNeuralNetwork}
\begin{figure}
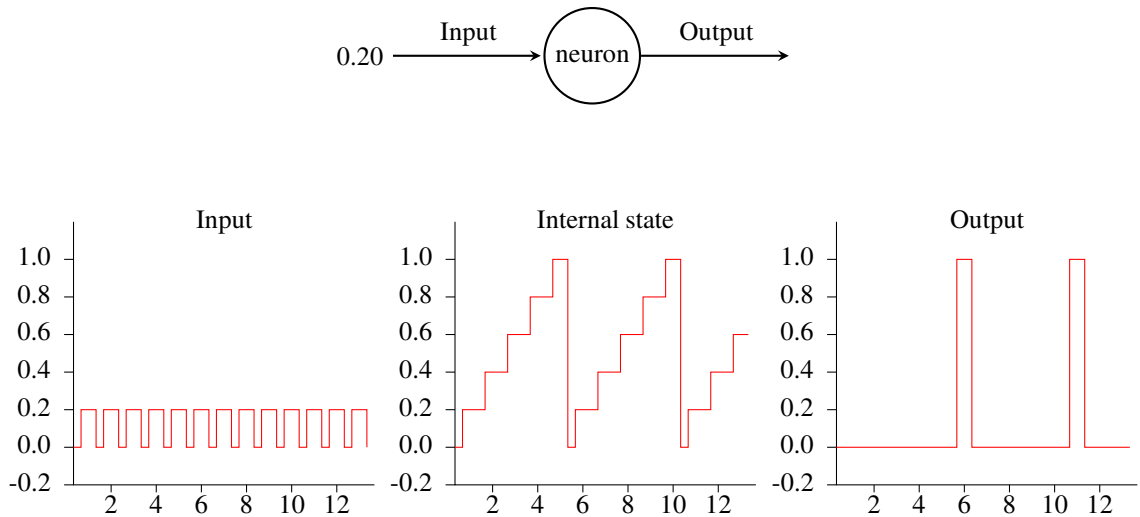

    \centering



    \caption{Activation of a spiking neuron}
    \label{figSpikingNeuron}
\end{figure}

\begin{figure}
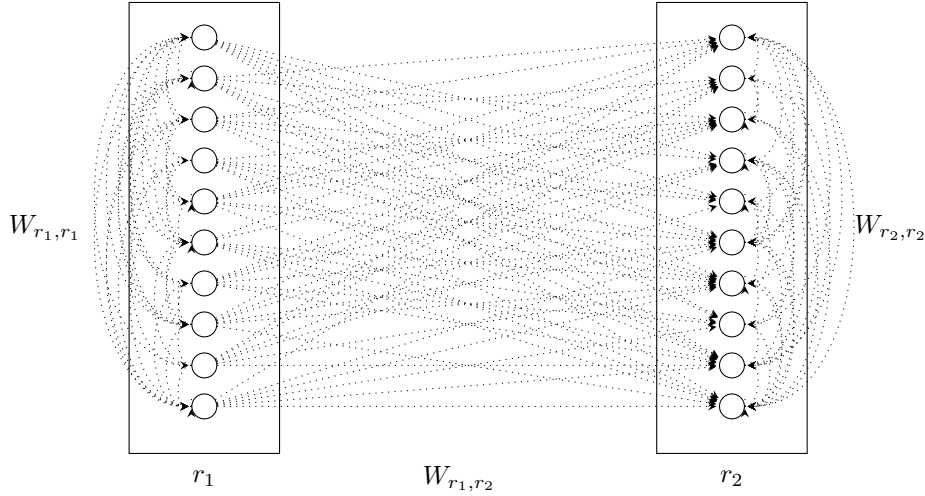

    \centering
    %
         
    \caption{Simple assemblage of two clusters}
    \label{figSimpleNetwork}
\end{figure}

\begin{figure}
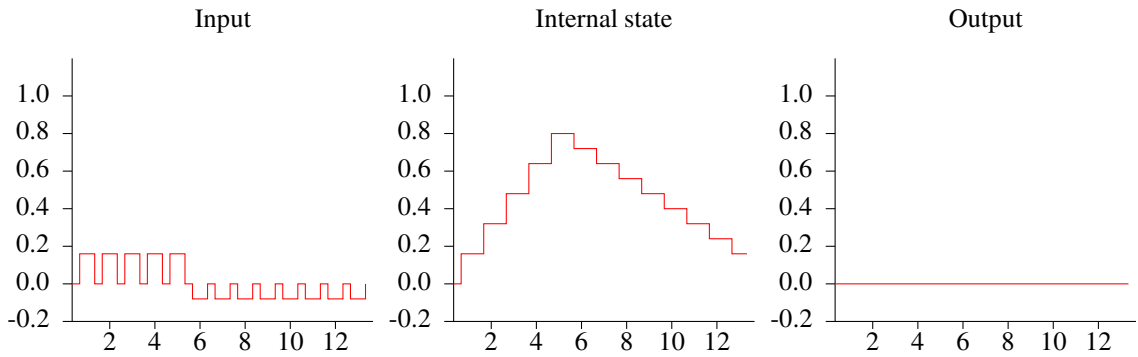

    \centering

    %
    \caption{Winner-take-all mechanism}
    \label{figWinnerTakeAll}
\end{figure} 

    This section presents an assemblage of spiking neural networks as a framework to implement prime attractors, registers, one-shot symbolic memory, and hash tables.
    In later sections, the extension of the framework accounts for symbolic computers.

    We use a simple model of spiking neurons. 
    It uses discrete time: at every time step, every neuron integrates its input and produces an output.
    Figure \ref{figSpikingNeuron} illustrates how the sum of the inputs relates to the output.
    We maintain an internal state for each neuron incremented at each time step by the weight of the synapse of each incoming neuron that spikes.
    When the state reaches a threshold of 1, the neuron spikes and at the next time step, adds the weight of the connecting synapse to the outgoing neurons' state, and resets its internal state to zero. 
    In figure \ref{figSpikingNeuron} we can see on the input graph that the neuron receives each time step input of 0.2 units which gets added to the internal state for which we show the graph as well.
    After each five-time step, the neuron spikes, its internal state is reset to zero, and we can see the output on its graph spike to 1 for a one-time step, remaining at zero the rest of the time.

    To implemented this we introduce a spike function $S$ defined here on a scalar, which essentially extracts the spikes.
    Its definition is also extended to a vector by applying it component-wise.
    \begin{equation}
        S(x) = \begin{cases}
            1,& \text{if } x\geq 1\\
            0,& \text{otherwise}
        \end{cases}
    \end{equation}
    We also define the clamping function $ C_a^b $ below on a scalar $x$. Its definition is extended to a vector by applying it component-wise.
    \begin{equation}
        C_a^b(x) = \begin{cases}
            a,& \text{if } x\leq a\\
            b,& \text{if } x\geq b\\
            x,& \text{otherwise}
            \end{cases}
    \end{equation}
    The following equation partly represents the assemblage of feedback spiking neural networks.
    \begin{equation}
        X_{i, n+1} = (1 - l_i)(C_0^1(X_{i, n}) - S(X_{i, n})) + \sum_j W_{i, j}S(X_{j, n}) \label{spikingRNEq2}
    \end{equation}
    First, note that the network architecture is not layered as usually done in deep learning but instead forms a graph.
    So instead of having a stack of layers, we have a directed graph of clusters with cycles.
    Further note, there are no biases, as they do not reveal to be necessary for our purpose.
    The variable $n$ represents the time step. The expression $X_{i, n}$ represents a vector holding the internal state of the neurons forming the cluster $i$.
    A cluster is a group of neurons acting together.
    The expression $W_{i, j}$ is the weight matrix linking cluster $i$ to cluster $j$.
    Note that $W_{i,j}$ is a sparse matrix when not the zero matrix altogether.
    This restriction represents the brain's limited number of neuron-to-neuron connections. 
    Also, the variable $l_i$ represents the leaking factor of the neuron to model the leaking of a spiking neuron. 
    The leaking is not required in most clusters, but its usage is to prevent noise from accumulating and causing some unwanted neurons to fire in panel clusters seen in section \ref{PanelClusters}.
    The expression $C_0^1(X_n) - S(X_n)$ represents the state of the neurons after the effect of spiking if this occurs.
    Furthermore, the multiplication by $(1 - l_i)$ represents the resulting state after leaking.
    Note that most of the $W_{i, j}$ are the zero matrices; only if cluster $i$ connects to cluster $j$, it is not zero.
    Figure \ref{figSimpleNetwork} shows a simple assemblage with two clusters $r_1$ and $r_2$, each containing 10 neurons for illustrative purposes an their connections. 
    The state of the neurons of cluster $r_1$ at time step $n$ is $X_{r_1, n}$ and of cluster $r_2$, $X_{r_2, n}$.
    The connections are $W_{r_1, r_1}$, $W_{r_1, r_2}$, $W_{r_2, r_2}$, where each is composed of a number of links connecting neurons of the input cluster to neurons of the output cluster (which might be the same).  
    There is no $W_{r_2, r_1}$ present on figure \ref{figSimpleNetwork} to mean that it is the zero matrix.
    In practice, in this work, each cluster is sometimes connected to itself, has incoming connections from at most two clusters, and has outgoing connections to at most two clusters.
    This limit, however, does not include control connections seen in section \ref{SymbolicFramework};
    
    This simplified version of a spiking neuron is sufficient to implement a winner-take-all mechanism\cite{lynch2019winnertakeall}.
    We illustrate in figure \ref{figWinnerTakeAll} how it works in our implementation with two neurons $A$ and $B$.
    When $A$ spikes, it outputs 1 unit to itself and inhibits $B$ by 0.24 units, similarly for $B$.
    Suppose $A$ and $B$ internal states start at zero; $A$ receives an outside input of 0.20 units at each step; while $B$ receives an outside input of 0.16.
    Then, both neurons enter a race where their internal state will increase until one reaches 1 first, which is $A$.
    At this point, $A$ adds to its input its output and inhibits the neuron $B$, with the effect that $A$ spikes continuously beyond this point, and $B$'s internal state decreases and never spikes.
    We use advantageously this effect to implement prime attractors in the next section \ref{PrimeAttractor}.

    
    Equation \ref{spikingRNEq2} also does not include a second-degree network that is used to implement hash tables and is introduced in section \ref{Hashtable}.

    These are the principal characteristics of our framework.

    Next we look at how symbols are implemented.

\subsection{Symbols in Distributed Representations} \label{DistributedRepresentation}

    We suggest that predefined sparse activation patterns represent symbols in a distributed representation.
    That is, we assign to each symbol a sparse random activation pattern to represent it.
    The only characteristic of the random activation patterns is their coverage that is the ratio of firing neurons. 
    We chose this number to be the same for all the random activation patterns.

    Given a list of symbols, we assign each one to a sparse random pattern of activation of a given coverage to represent it.
    We observe these patterns of activation:
    \begin{itemize}
        \item Like symbols, they have no intrinsic information because their activation pattern is random.
        \item Like symbols, they have no internal structure for the same reason.
        \item The part in common between two random patterns is due to chance and does not carry any meaning or information other than being part of the two patterns.
        \item Any substantial preselected part represents the whole; that is, if one recognized the part, one knows which pattern it is. 
        By substantial part, we mean of size given the sparsity of the patterns such that they are enough neurons active so that by chance, two different patterns taken from the list cannot have, in practice, these same neurons active. Thus, these active neurons uniquely identify the patterns.
        This property makes a register switch box, section \ref{RegisterSwitchbox}, physiologically plausible.
        \item Therefore, in some sense, they are atomic, like symbols despite being a distributed representation.
    \end{itemize}

    This is how symbols are represented in this framework.
    We now proceed with the implementation of symbolic components.

\subsection{Prime attractors and registers} \label{PrimeAttractor}

    Again, given a list of symbols, we assign each one to a sparse random pattern of activation of a given coverage to represent it.
    A prime attractor converges toward the random pattern assigned to a symbol, and a register is the network that holds the prime attractor.
    They are used to implement the one-shot symbolic memory, the hash table and the register switch box.
    \emph{Prime attractors} are fix-point attractors generated out of sparse saturated random noise representing symbols.

    We define a \emph{register} as a cluster with a feedback connection trained to recognize prime attractors.
    The role of the register is first to converge towards one of learned prime attractors by filtering out the noise from the input, then to sustain the prime attractor it has converged to.
    We say it recalls the prime attractor.
    The prime attractors are also fully saturated. 
    In practice, their components will be only zeros or ones, where one is the activation threshold.
    The zero or one values help in making the prime attractor self-sustaining, which is essential to have our register network act as a register in a computer holding a value.
    Limiting the components of a prime attractor to zero or one also means sets can model them. 
    We interchangeably treat prime attractors as sets or as vectors.
    Prime attractors have many remarkable properties: 
    \begin{itemize}
        \item They are initially devoid of information or substructure because they are random. 
        The information associated with the random attractors is given to them when they are bound to other attractors or patterns, or conversely, other attractors are bound to them. 
        Prior to that, they are interchangeable. 
        \item The only value characterizing an attractor would be its coverage, the ratio of 1's. 
        In practice, all the prime attractors recognized by a network have the same coverage, making it a property of the network.
        We want the coverage to be small because it does, as appendix \ref{ShorttimeMemoryCapacity} shows, affect the capacity of the memory.
        \item Another advantage of prime attractors is that we believe the neural network generating them is not subject to spurious attractors\cite{Hopfield:1982:Proc-Natl-Acad-Sci-U-S-A:6953413}
        \cite{journals/nn/RobinsM04} as there are no sub-attractors in common that could be recombined to create them.
        \item They are used to implement one-shot symbolic memory, section \ref{OneShotSymbolicMemory} the hash table, section \ref{Hashtable} and the register switch box, section \ref{RegisterSwitchbox}.
    \end{itemize}

    In this work, a register $r$ has its self-connection $W_{r, r}$ trained to recognize a set of $e$ prime attractors $A_i$'s, that is, random pattern of 0's and 1's with all the same coverage $c$,
    using the following equation.
\begin{equation}
    A_i = C_0^1(\frac{W_{r,r}A_i + \alpha \overrightarrow{1}}{1 + \alpha}) \label{attractorEq}
\end{equation}
    In this way, one trains the weight matrix $W_{r,r}$ such that $W_{r,r} A_i$ produces an output of at least 1 on components where $A_i$ is 1 but produces an inhibition response of $-\alpha$ or less on components where $A_i$ is 0. 

    In figure \ref{figSymbolicMemory}, there are two registers $r_1$ and $r_2$ with the feedback connections $W_{r_1, r_1}$ and $W_{r_2, r_2}$ respectively.
    For figurative purpose, each register contains 10 neurons with 3 active (the darken ones) representing its prime attractor of coverage $\frac{3}{10}$.

    The inhibition response is crucial because we want the prime attractor to inhibit the competing prime attractors as it forms in conjunction with any noise, giving its noise-filtering capability to the attractor.
    This uses the winner-take-all mechanism\cite{lynch2019winnertakeall} illustrated with figure \ref{figWinnerTakeAll} at the neuron level but pattern-wide instead.
    Note that this equation does not work for the equivalent non-spiking neural network (see appendix \ref{spikingvsnonspiking}) as the inhibition response is too strong and overwhelms the network.
    This inhibition response is not as prevalent in a spiking network because of the winner-take-all mechanism\cite{lynch2019winnertakeall} that the equation \ref{attractorEq} naturally implements.
    Only the inhibition given by the first neurons to spike is taken into account if they correspond to a given prime attractor. 
    The inhibition signals sent by the first neurons to spike inhibit the neurons of the competing prime attractors before they send their inhibition signal.
    The network then reinforces the given prime attractor and recalls it.

    Moreover, in our implementation, we train the attractors directly and do not use a spiking network, facilitating the learning. 
    Note that the learning is done on a single layer and thus does not require backpropagation, making it learnable using only local rules, therefore biologically plausible.

    With this building block that is the register recognizing prime attractors, we proceed to implement the one-shot symbolic memory and the hash table, which are main components of our symbolic computer.

\subsection{One-shot symbolic memory} \label{OneShotSymbolicMemory}

\begin{figure}
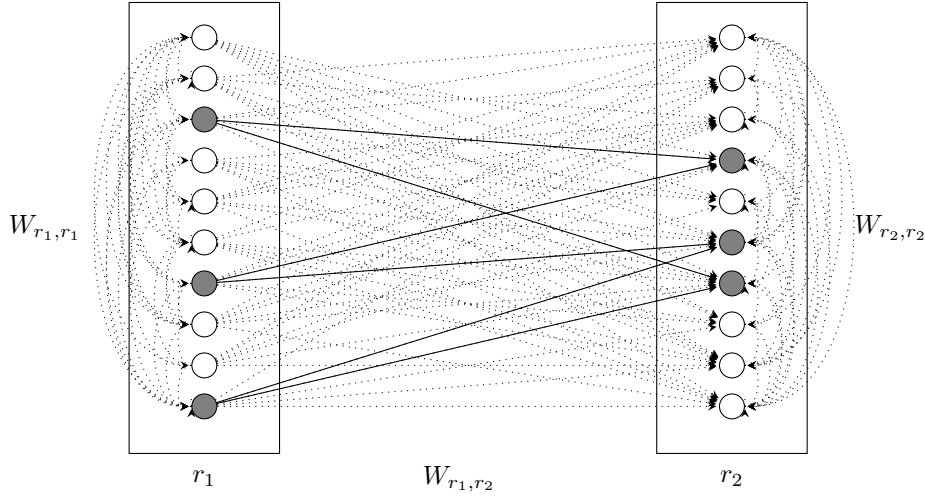

    \centering
         
    \caption{Symbolic Memory during binding/unbinding}
    \label{figSymbolicMemory}
\end{figure}

\begin{figure}
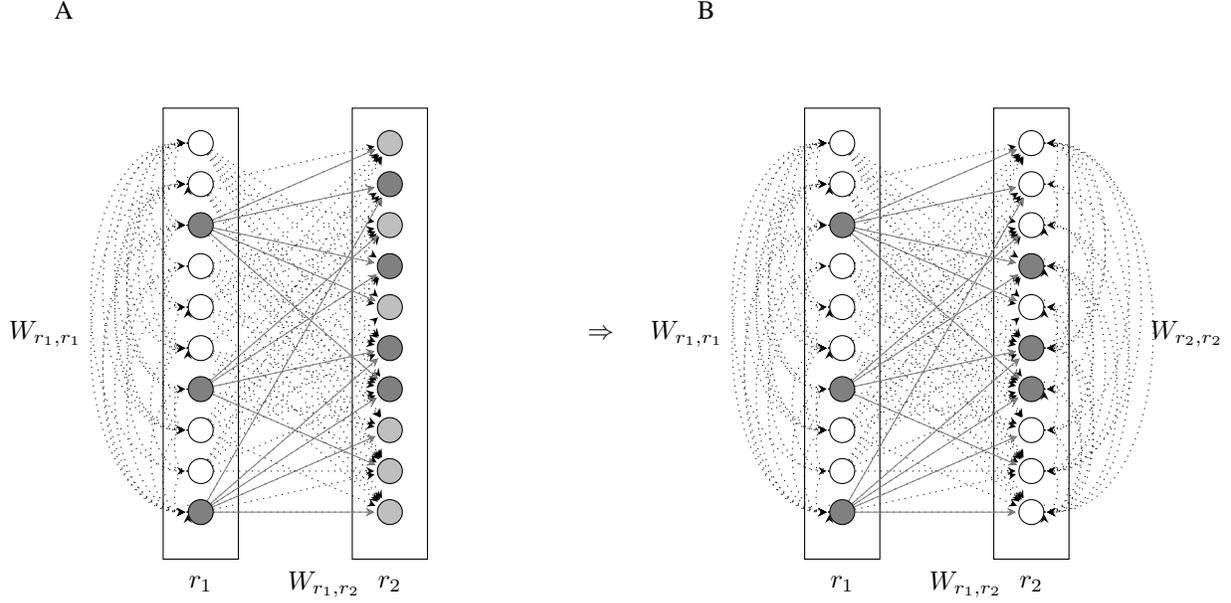

    \centering
    %
         
    \caption{Symbolic Memory: Role of feedback connection $W_{r_2,r_2}$ during recall}
    \label{figSymbolicMemoryRoleOfPrimeAttractor}
\end{figure}

    The ability to store a value in a prime attractor is a crucial functionality enabling a symbolic computer.
    The value that can be stored is a prime attractor as well; that is, we associate through a network a prime attractor to another prime attractor.
    We present in this section this one-shot symbolic memory, which resides in the synapses connecting two registers and takes advantage of the sparsity and the randomness of activation of the prime attractors. 
    
    Essentially, we propose a Hebbian learning scheme \cite{hebb}, which relies on the noise-filtering capacity of the register to compensate for the noise introduced by the imprecision of the learning. 

    The excitatory synapses linking an input register to an output register are chosen randomly and set to zero.
    We refer to them as the \emph{one-shot symbolic memory} connection.

    We illustrate the mechanism in figure \ref{figSymbolicMemory}. 
    \begin{itemize}
        \item $r_1$ is a register of coverage $c_1$ and $r_2$ is a register of coverage $c_2$.
        \item $W_{r_1, r_2}$ is the one-shot symbolic memory connection linking $r_1$ to $r_2$.
        \item $W_{r_2, r_2}$ is the self connection of $r_2$
        \item For illustrative purpose each register has 10 neuron represented by circles (in practice for the symbolic computer we implemented we use registers of 10000 neurons).
        \item The darken circles represents the prime attractor active on each register.
        \item The arrows between the neuron of $r_1$ and $r_2$ represents the synapes connecting them.
        \item The darken arrows represents the synapses linking a firing neuron to a firing neuron and those in a dotted line the others.
    \end{itemize}
    The principle is that using one-shot learning, 
    given an input prime attractor in the input register $r_1$ and an output prime attractor in the output register $r_2$,
    the network learns with the connecting synapses (those in black) an approximation of the desired output prime attractor.
    Moreover, the mechanism relies on $W_{r_2, r_2}$ to filter out the noise and recover the output prime attractor during recall.
    This is illustrated in figure \ref{figSymbolicMemoryRoleOfPrimeAttractor}, where we show the content of $r_2$ without $W_{r_2,r_2}$ on side A and with it on side B.
    On side A, the frequency of the activation of the neurons is given by the shade of gray of the neuron.
    
    The connection learns the approximation with a simple Hebbian-like rule on the connecting synapses summarized by "neurons wire together if they fire together" \cite{lowel:science92}.
    In our version, 
    we set the connecting synapses linking a firing neuron of the input prime attractor with a firing neuron of the output prime attractor to a nominal value $a$.
    That is, the darken arrows in figure \ref{figSymbolicMemory} are given the weight $a$.
    We chose the value $a$ such that each output firing neuron receives a total potential of a fraction of the activation threshold, so it fires after a few steps.
    For instance, if we chose $a$ such that the total received potential for an active neuron is 0.2 on average, the neuron should fire during a recall after 5 or 6 cycles.
    We do not want to reach full activation in one step because we must distinguish the noise from the signal we wish to recover.
    For instance, if the signal we wish to recover makes its neurons spike after 6 cycles, but the other noisy neurons take 7 cycles, 
    the noisy neurons never fire because the neurons of the signal inhibit them. 
    Given that the input prime attractor and the output prime attractor have coverage $c_1$ and $c_2$ respectively, 
    then the ratio of synapses involved in the learning is only $c_1 c_2$, assuming $c_1$ and $c_2$ are small. 
    Of course, as more prime attractors are bound, the various learnings interfere with each other.
    The one-shot symbolic memory connection given an input prime attractor produces a noisy signal containing the bound output prime attractor.
    We then expect the output register to recover the output prime attractor, as illustrated in figure \ref{figSymbolicMemoryRoleOfPrimeAttractor}.
    The key idea we discuss in appendix \ref{ShorttimeMemoryCapacity} is that because the prime attractors are random patterns, the noise is uniformly distributed and does not accumulate anywhere.
    This idea is not entirely accurate as the noise accumulates around the synapses of the prime attractors that are bound to multiple different prime attractors.
    The uniformity of the noise facilitates the noise filtering by the output register to filter out the noise.
    Assuming each prime attractor is bound only once, the capacity is naively lower than $\frac{1}{c_1 c_2}$ as it would nearly saturate the synapses, 
    making the signal supporting the value bound indistinguishable from noise.

    We can actively "forget" a binding by the same way we do binding, that is, 
    by making an opposite Hebbian rule and resetting to zero the synapses connecting a firing neuron to a firing neuron. 
    That is, the darken arrows in figure \ref{figSymbolicMemory} are given the weight 0.
    Actively forgetting a binding is not the same as never learning it, as the act of forgetting damage the other bindings slightly. 
    It would be minimally disruptive for other learned bindings since only a minimal amount of synapses, 
    a ratio of $c_1 c_2$ is affected, assuming $c_1$ and $c_2$ are relatively small. 
    We did not address this issue in this work as it did not prove to be a problem in the examples we did, but one can presume that longer-term memories that use this mechanism would need to be refreshed periodically.
    Forgetting is important because it restores the capacity of the memory.

    In appendix \ref{ShorttimeMemoryCapacity}, we derive that:
    \begin{equation}
        n \leq \frac{1 - s c_1 - t c_1 c_2}{\gamma c_1 c_2} - \frac{m}{c_2} - \frac{1}{c_1}
    \end{equation}
    Where:
    \begin{itemize}
        \item variable $n$ is the capacity, 
        \item variable $s$ is the number of times, after being bound, a prime attractor in the memory has been unbound to another attractor,
        \item variable $t$ is the number of unboundings since the oldest bound prime attractor,
        \item variable $\gamma$ is the signal-noise ratio, which we also discuss the values in appendix \ref{ShorttimeMemoryCapacity}, 
        \item and variable $m$ is the maximum number of times we bind the same symbol to other symbols.
    \end{itemize}

    This limit on $m$ is a significant constraint in this working memory.
    Nevertheless, we may palliate by using proxy symbols to denote the same symbol for popular symbols.
    This proxy technic also reduces the value of $s$.
    It is also apparent that the variable $s$ and $t$ should not be allowed to grow too much by refreshing the memories periodically.

    In appendix \ref{ShorttimeMemoryCapacity}, we have an example with $\gamma = 5$, $m = 20$, $c_1 = c_2 \frac{30}{10000}$, $s = 30$ and $t = 10000$, which gives $n < 11222$. 
    
    It is also possible to create a default value. 
    One implementation involves another set of synapses going from any input pattern to the default prime attractor with half the intensity used by the standard binding mechanism.
    Thus, the network should induce the default value with half the intensity on any given input prime attractor. 
    During a recall, if no prime attractor is bound to the prime attractor of the input register, this half intensity should be higher than noise, and the output register should recover the default value.
    Suppose a prime attractor is bound to the prime attractor of the input register. The bound prime attractor should receive full intensity, and the output register should select it over the default value.
    This mechanism, however, diminishes the capacity of the memory.

    Now that we've shown that prime attractors can act as variables, we next extend the one-shot symbol memory mechanism to handle hash tables.

\subsection{Hash table} \label{Hashtable}

\begin{figure}
    \centering
         
    \caption{Hashtable}
    \label{figHashtable}
\end{figure}
 
    Hash tables are a valuable component to build programs. 
    This implementation using prime attractors capitalize on the sparsity and the randomness of their activation pattern.

    Hash tables are not required to build a symbolic computer. It is not clear this implementation is physiologically plausible, but they are interesting nonetheless.

    They assume a second-degree network, that is, a type of neuron that we made up with dendritic branches that compute the \emph{and} function on their sole two synapses.
    The neuron fires only if the two synapses of any of its dendritic branches are excited. 

    In figure \ref{figHashtable}, we illustrate the hash table mechanism. The column of squares on the left of the \emph{hash} register corresponds to the dendritic branches.
    Each dendritic branch connects to its right to the $X_{\text{hash}}[k]$ neuron it belongs to and to the left to a $X_{\text{table}}[i]$ neuron and a $X_{\text{key}}[j]$ neuron.
    The dark squares correspond to the dendritic branches having their two synapses excited which trigger their $X_{\text{hash}}[k]$ neuron. 
    The darken arrows coming out of the \emph{table} register and the \emph{key} register represent the synapses of these activated dendritic branches.

    Given \emph{table}, \emph{key} and \emph{value} registers with coverage $c$ and a cluster \emph{hash}.
    We define a second order network, taking as input the spiking output of the \emph{table} register, the $S(X_{\text{table}})[i]$'s and of the \emph{key} register, the $S(X_{\text{key}})[j]$'s, with the following equation with most $\omega_{ijk}$ equal to 0 
    and the others selected at random equal to 1 with the property, given $k$, that $|\{ (i, j) | \omega_{ijk} \neq 0 \}| = \lfloor\frac{1}{c}\rfloor$.
\begin{equation}
    X_{\text{hash}}[k] = \sum \omega_{ijk} S(X_{\text{table}})[i] S(X_{\text{key}})[j] \label{hashTableEq}
\end{equation}
    The idea is that the \emph{hash} cluster looks like a random pattern of coverage approximately $c$, given that the \emph{table} register holds a random pattern of coverage $c$, the \emph{key} register holds a random pattern of coverage $c$, and the $\omega_{ijk}$'s are randomly selected as described. 
    The goal is to use the \emph{hash} cluster as input to a one-shot symbolic memory connection instead of an input register, 
    allowing the attractor held by the \emph{value} register to be bound to the attractors of the \emph{table} register and the \emph{key} register.
    This is illustrated in figure \ref{figHashtable}.
    
    Note that this second-order network does not require any learning.  We just hardwired it like this at random. 
    Now what is left to do is bind the value of the \emph{value} register to the value of the \emph{hash} cluster with our one-shot symbolic memory, the \emph{hash} cluster replacing the input register.
    This mechanism effectively implements a hash table. 
    The hash table benefits significantly from a default value despite reducing the capacity of the hash table since the program does not need to know which entries have values.

    Next, we take a look at numbers in relation to the brain.

\subsection{Relating to the brain} \label{RelatingToTheBrain}

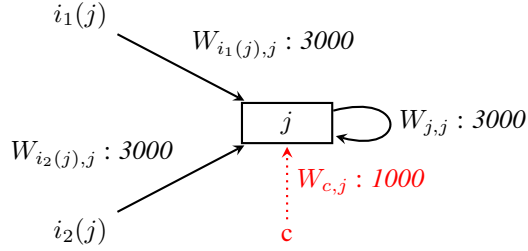
\begin{figure}
    \centering
    
    \begin{tikzpicture}[->,>=stealth,shorten >=1pt,auto,node distance=2.8cm,thick]
 
        \tikzstyle register=[draw, minimum width=1.2cm]
        \tikzstyle control=[draw, minimum width=1.2cm]

        \node[register] (j)     {$j$};
        \node           (anchori) [left=2cm of j] {};
        \node (i1)      [above=1cm of anchori] {$i_1(j)$};
        \node (i2)      [below=1cm of anchori] {$i_2(j)$};
        \node [red] (c)       [below=1cm of j] {c};

        \path   (j)    edge [loop right] node {$W_{j,j}$ : \emph{3000}} (j)
                (i1)    edge  node {$W_{i_1(j), j}$ : \emph{3000}} (j)
                (i2)    edge  node {$W_{i_2(j), j}$ : \emph{3000}} (j);

        \path[red, dotted] (c)  edge node[right] {$W_{c, j}$ : \emph{1000}} (j);
         
    \end{tikzpicture}
    \caption{Limit on the number of connections}
    \label{figLimit}
\end{figure} 
    
    Now that we have our components to build a symbolic computer. 
    Let us look at numbers and discuss the limitations of the brain our framework should reflect and their implications.

    The human brain has no shortage of neurons, so the number of neurons is not a limiting factor.
    However, for computational reasons, we keep the network at a manageable size in the examples.  
    Each neuron has a limited number of connections.
    On average, a neuron connects to $10000$ neurons in the human brain. 
    This limit means that for a given cluster $j$, the $W_{i, j}$ of all the clusters $i$ connected to it can have at most $10000$ connections per neuron.
    In this work, we limit a cluster to three incoming clusters $i$ including itself with each $3000$ connections per neuron.
    This limit of 3 does not affect control connections seen in section \ref{SymbolicFramework}.
    That is, as shown in figure \ref{figLimit}:
    cluster $j$ has itself for a self-connection $W_{j, j}$ and two other input cluster $i_1(j)$ and $i_2(j)$ with connection matrices $W_{i_1(j), j}$ and $W_{i_2(j), j}$ respectively. 
    Each of $W_{j, j}$, $W_{i_1(j), j}$ and $W_{i_2(j), j}$ is a sparse matrix connection where each input neuron connects to at most $3000$ output neurons (see figure \ref{figLimit}).
    We save the $1000$ remaining connections per neuron for the control connections seen in section \ref{SymbolicFramework}.

    This limit on the number of connections per neuron directly impacts the minimal size of an attractor since the neurons participating in the attractor must provide feedback to themselves.
    Suppose a sparse random pattern of activation with a coverage $c$ (defined here as the ratio of active neurons in the attractor),
    and suppose that each neuron in a cluster connects randomly to $\kappa$ neurons uniformly,
    then each neuron receives feedback from $c \kappa$ neurons of the attractor. 
    If we want the feedback from at least $\theta$ neurons, we get the following inequation:
    \begin{equation}
        c \geq \theta / \kappa
    \end{equation}        
    In practice, we want feedback from more than one neuron. Appendix \ref{ShorttimeMemoryCapacity} discusses the values for $\theta$.
    If one uses ten neurons for $\theta$, this means that the minimum coverage is $\frac{1}{300}$ if each neuron connects to $3000$ neurons.
    A minimal coverage $c$ is essential: the smaller the attractor, the fewer resources it requires to bind it, 
    and the higher the capacity of the memory (as shown in appendix \ref{ShorttimeMemoryCapacity}).

    How many prime attractors can a register network support? It is an important question.
    We have empirical evidence. The limit of $3000$ connections on the self-connection seems significant.
    In an experiment, with a cluster of $3000$ neurons fully connected, we produced $25000$ prime attractors of 6 neurons that we were successfully bound in sequence and recalled.
    However, these prime attractors are not suitable for use with the hash table.
    For primes attractors suitable for the hash table in another experiment, we use a cluster of $10000$ neurons self-connected, connected to $3000$ neurons each.  
    In this case, we did generate $5000$ attractors that did about $5000$ bindings.

    These limits constrain the symbolic computer we propose building in the next section.
    However, we believe that symbolic computation might capture an aspect lacking in today's deep learning networks and complement them advantageously even in small doses.
    These limits might also explain why the human brain is so feeble at symbolic computations despite being a powerhouse of computation with its 100 billion neurons.
    
    Now that we have defined the main components required in the construction of a symbolic computer, but before we proceed to build it, we flesh out the symbolic framework we use to build it.
    
\section{Symbolic framework} \label{SymbolicFramework}

    The assemblage of feedback spiking neural network introduced by equation \ref{spikingRNEq2} needs to be augmented to serve as a framework to build a symbolic computer.
    Notably, we need control connections to trigger the binding/unbinding of the one-shot symbol memory.
    A discussion about the encoding used by the clusters follows with one about the various type of connections linking the clusters together.
    The section concludes by showing how to build a register switch box, which is at the core of the symbolic computer we build in the following section. 

\subsection{Upgrade of the equation for the symbolic framework} \label{SymbolicFrameworkUpgrade}
    Equation \ref{spikingRNEq2} represents only a partial representation of the symbolic framework since there is also another type of connection it does not account for, the control connection.
    One type of control connection is the inhibition connection.
    The inhibition connection sends a strong inhibition signal to each neuron of the cluster it controls when active.
    This type of control connection can be and is represented by $W_{i, j}$.
    There are, however, two other types of control connection: the \emph{binding} connection and the \emph{unbinding} connection.
    We propose a mechanism similar to the inhibition of all the neurons of a cluster.
    The presynaptic neurons of the cluster receive a \emph{binding} message, which triggers the \emph{binding}.
    Similarly, the presynaptic neurons of the cluster receive an \emph{unbinding} message, which triggers the \emph{unbinding}.
    That is, the control connection signals to each presynaptic neuron of the cluster $i$ it controls to engage in a binding and unbinding behavior, respectively, which will modify the $W_{i,j}$ as shown in section \ref{OneShotSymbolicMemory} given that the connection between cluster $i$ and cluster $j$ is of the one-shot symbolic memory kind.
    We could as well have chosen the postsynaptic neurons to receive the message. 
    From an artificial perspective, this choice is arbitrary. 
    It does not indicate whether it would be the presynaptic or the postsynaptic neurons that receive the message if such a mechanism exists in the brain. 
    These control connections enable the system to act as a conventional computer.

    The \emph{binding} connection and the \emph{unbinding} connection does not affect equation \ref{spikingRNEq2} directly.
    Rather it affects the variable $W_{i,j}$ which gets updated if there is a binding or an unbinding on it.

    Equation \ref{spikingRNEq2}, however, needs to get updated to account for hash tables.
    \begin{equation} \label{spikingRNEq3}
        X_{i, n+1} = (1 - l_i)(C_0^1(X_{i, n}) - S(X_{i, n})) + \sum_j W_{i, j}S(X_{j, n}) + H_i(X_{a_i, n}, X_{b_i, n})
    \end{equation}
    The function $H_i$ is either the zero function or represents equation \ref{hashTableEq}, where cluster $i$ is the \emph{hash} cluster, cluster $a_i$ is the \emph{table} register and cluster $b_i$ is the \emph{key} register.

    Equation \ref{spikingRNEq3} represents the framework for building our symbolic computer.
    However, it does not say anything about the encoding used by each cluster.

\subsection{Encodings} \label{Encodings} \label{PanelClusters}

    In our symbolic framework, every cluster is one of the following two types depending on the encodings for its firing pattern.
    \begin{description}[align=left,labelwidth=2cm]
        \item [Symbol cluster]
        A \emph{symbol cluster} uses a random sparse pattern encoding. 
        The cluster is associated with several sparse random patterns of activation of a given coverage $c$ representing symbols.
        The cluster in a stable state is then assumed to converge towards one of the patterns.
        This is the encoding used by registers for prime attractors.
        These clusters hold symbols.

        \item [Panel cluster]
        A \emph{panel cluster} uses a per neuron encoding.
        With this encoding, every neuron has a designation. 
        A per neuron encoding is unrealistic for the brain, in which case we could consider group neurons acting like one instead to provide redundancy.
        The use of per neuron encoding here is for simplicity. 
    \end{description}

\subsection{Connections} \label{Connections}
 
Given a connection between two clusters, the type of the clusters may serve as a classifier.
According to that classification, there are four types of connections that our symbolic computer uses:
\begin{itemize}
    \item symbol cluster to symbol cluster,
    \item symbol cluster to panel cluster,
    \item panel cluster to symbol cluster,
    \item panel cluster to panel cluster.
\end{itemize}

\subsubsection{Symbol cluster to symbol cluster} 
    This type of connection corresponds to the self-connection of a register defining a set of prime attractors.
    It also corresponds to the \emph{one-shot symbolic memory connection} used in section \ref{OneShotSymbolicMemory}.
    Based on the self-connection training equation, equation \ref{attractorEq}, the definition of a \emph{mapping connection} between two clusters uses the following training equation:
    \begin{equation} \label{mappingEquation}
        B_k = C_0^1(\frac{W_{i,j}A_k + \alpha \overrightarrow{1}}{1 + \alpha})
    \end{equation}
    The training of $W_{i,j}$ uses this equation so that each given input firing pattern $A_k$ of cluster $i$ maps to the given firing pattern $B_k$ of cluster $j$.
    Equation \ref{mappingEquation} is a generalization of equation \ref{attractorEq} on two clusters with potentially different sets of firing patterns.
    If the mapping connection maps each pattern to the same firing pattern, and those patterns cover all the patterns supported by the two clusters, we call it an \emph{assignation connection}.
    The self-connection of a register is an instance of an assignation connection, except it serves the purpose of reinforcing the pattern.
    However, this definition assumes the two clusters accept the same random sparse patterns, which is unrealistic for the brain.
    The two clusters, say $n_1$ and $n_2$, would recognize a different set of random sparse patterns in a natural setting.
    Nevertheless, in that case, we can say that if there is a bijunction mapping the random sparse patterns of $n_1$ and the random sparse patterns of $n_2$ to the same set of symbols, called $P$.
    Then, we define an \emph{assignation connection} between the two clusters, as a mapping connection that maps for each symbol $a$ of $P$, the activation pattern associated with $a$ in the input cluster to the activation pattern associated with $a$ in the output cluster.

    \subsubsection{Symbol cluster to panel cluster} 
    This type of connection is a random sparse pattern recognizer, we call it a \emph{recognizer connection}. 
    It detects random sparse patterns of the input cluster and fires specific neurons in the output cluster if they are present.
    The assignation of a nominal weight of $a$ to every synapse connecting a neuron of a random sparse pattern to detect to its output neuron gives an implementation. 
    The choice of the value of $a$ is such that the output cluster fires when the pattern is active on the input cluster but does not fire when another pattern is active on the input cluster.
    Given that the input clusters accept patterns of coverage $c$, and a pattern is active on the input cluster, a neuron of the output cluster meant to detect another pattern might receive inputs from about $\frac{1}{c}$ neurons of the input pattern instead of all of them.
    The input is insufficient to fire initially for a reasonable value of $a$. 
    However, to prevent the input from accumulating and eventually causing the neuron to fire, we give the output cluster a decisive leaking factor.  The effect for the neuron receiving the input, if it does not fire immediately, is to forget about it and thus only fires when the correct input pattern is present.

    \subsubsection{Panel cluster to symbol cluster}
    \emph{Control connections} use this type of connection to broadcast a message to all the neurons of a symbol cluster.  
    A neuron, when firing, can send a strong inhibition signal to all the neurons of a cluster.
    Another neuron, when firing, can send a solid binding signal to all the neurons of a cluster, similarly, for the unbinding.
    For instance, a \emph{control cluster} which is a panel cluster where every neuron, when firing, broadcast to a cluster either a strong inhibition signal, a binding signal, or an unbinding signal.

    One issue with the broadcast encoding is that it does not respect the constraint of section \ref{RelatingToTheBrain} on the limit of connections between neurons.
    However, this can quickly be resolved by using multiple layers instead of connecting directly.
    For instance, each neuron of a panel cluster connects to a distinct set of neurons on an intermediate cluster which the neurons together broadcast to the output cluster.
    For this reason, control connections are not limited in their number per cluster.

    \subsubsection{Panel cluster to panel cluster}
    This connection represents a neuron triggering another neuron; we call it a \emph{per-neuron connection}.
    For instance, in the case of self-connection, a neuron can activate its successor. 
    Thus they may represent sequences of microinstructions where each neuron is a microinstruction that executes within a sequence.
    Each microinstruction neuron using that type of connection on a \emph{control cluster} can control which clusters receive an inhibition signal, doing binding or doing unbinding.

    There are also little tricks used with this type of connection on a self-connection.
    \begin{itemize}
        \item Have a neuron trigger itself, to have a neuron always firing which can be used by the other neurons to compensate the lack of bias in the framework equation \ref{spikingRNEq3};
        \item This ever firing neuron serves to implement a complement neuron to another neuron. 
        It fires whenever a given neuron does not fire and is silent otherwise.
        This mechanism may implement the complement of a pattern recognizer, that is, a neuron that fires whenever a given sparse random pattern is not present on a given cluster.
        \item For sequences of microinstruction, it is possible to implement a mechanism preventing the re-entry in the sequence as long as the first neuron of the sequence fires continuously.
        The first neuron of the sequence can fire a barrier neuron which prevents the double firing of the second neuron as long a the first neuron fires.
    \end{itemize}

    We have covered all the types of connections.
    Before presenting the whole architecture, the register switch box is the last component at the core of our symbolic computer.

\subsection{Register switch box} \label{RegisterSwitchbox}
    One key aspect of doing symbolic operations is the ability to transfer a symbol from one register to another.

    \begin{figure}
        \centering
        \begin{tikzpicture}[->,>=stealth,shorten >=1pt,auto,node distance=2.8cm,thick]
     
            \tikzstyle register=[draw, minimum width=1cm, line width=1pt]
            \tikzstyle cluster=[draw, minimum width=1cm, line width=1pt]
            \tikzstyle control=[draw, minimum width=1cm, line width=1pt, blue, rounded corners]
    
            \node[red]  (cA)                       {};
            \node[register] (r1A)     [left=0.7 cm of cA] {$r_1$};
            \node[register] (r2A)     [right=0.7cm of cA] {$r_2$};
            \node (A) [above left=4.0cm of cA] {A};
             
            \path   (r1A)    edge [loop left] node {$W_{r_1,r_1}$} (r1A)
                    (r2A)    edge [loop right] node {$W_{r_2,r_2}$} (r2A)
                    (r1A)    edge node {$W_{r_1,r_2}$} (r2A);
    
            \node[control]  (cB)          [right=8cm of cA] {$c$};
            \node[register] (r1B)     [left=1.4cm of cB] {$r_1$};
            \node[register] (r2B)     [right=1.4cm of cB] {$r_2$};
            \node[cluster]  (r1tor2B) [above=1.7cm of cB] {$c_1$};
            \node[cluster]  (r2tor1B) [below=1.7cm of cB] {$c_2$};
            \node (B) [above left=4.0cm of cB] {B};
             
            \path[red, dotted, loosely dotted, line width=2pt] (cB)  edge node {$W_{c,r_1}$} (r1B)
                            edge node {$W_{c,r_2}$} (r2B)
                            edge node {$W_{c,c_1}$} (r1tor2B)
                            edge node {$W_{c,c_2}$} (r2tor1B);
    
            \path   (r1B)    edge [loop left] node {$W_{r_1,r_1}$} (r1B)
                    (r2B)    edge [loop right] node {$W_{r_2,r_2}$} (r2B)
                    (r1B)    edge node {$W_{r_1,c_1}$} (r1tor2B)
                    (r1tor2B) edge node {$W_{c_1,r_2}$} (r2B)
                    (r2B)    edge node {$W_{r_2,c_2}$} (r2tor1B)
                    (r2tor1B) edge node {$W_{c_2,r_1}$} (r1B);
    
            \node (cluster0)   [above=3cm of B.east]      {};
            \node (cluster1)    [right=2cm of cluster0] {};
            \node (cluster2)    [right=0cm of cluster1] {Symbol Cluster};
            \node[cluster] (cluster3)    [left=0cm of cluster1] {name};
            \node (bcluster0) [below=0.3cm of cluster0] {};
            \node (bcluster1)    [right=2cm of bcluster0] {};
            \node (bcluster2)    [right=0cm of bcluster1] {Panel Cluster};
            \node[control] (bcluster3)    [left=0cm of bcluster1] {name};
            \node (assignation0) [below=0.3cm of bcluster0] {};
            \node (assignation1)    [right=2cm of assignation0] {};
            \node (assignation2)    [right=0cm of assignation1] {Assignation Connection};
            \node (control0) [below=0.3cm of assignation0] {};
            \node (control1)    [right=2cm of control0] {};
            \node (control2)    [right=0cm of control1] {Control Connection};
    
            \path[->,>=stealth,shorten >=1pt,auto,node distance=2.8cm,thick]   
            (assignation0)    edge              node {} (assignation1)
            (control0) edge [red, loosely dotted, line width=2pt]     node {} (control1);

        \end{tikzpicture}
        \caption{Transfering a value between two registers}
        \label{figTransferingTwoRegisters}
    \end{figure}
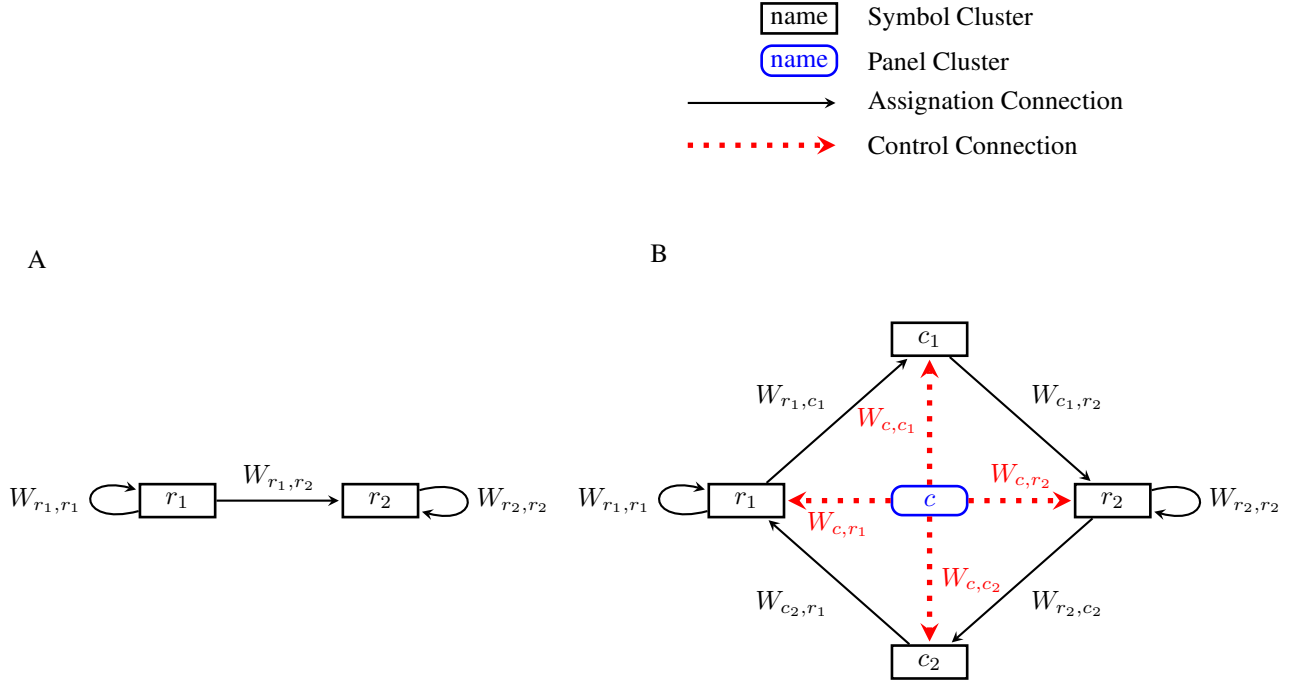 
    
    Note that figure \ref{figTransferingTwoRegisters} A is another representation of figure \ref{figSimpleNetwork} without displaying the neurons and the neuron-to-neuron connections.
    In figure \ref{figTransferingTwoRegisters} A, we see that if $r_2$ is cleared somehow, it will get the content of $r_1$ assuming the self-arrows on $r_1$ and $r_2$ and the arrow between $r_1$ and $r_2$ represent an assignation connection. 
    We need the means to clear $r_2$ to enable it to receive new content and to control the transfer of the content of $r_1$ to $r_2$. 
    In effect, we don't want the self-connection of $r_2$ to compete with the assignation connection between $r_1$ and $r_2$. 
    In figure \ref{figTransferingTwoRegisters} B, 
    we show a network with the capacity to move a value from $r_1$ to $r_2$ and then from $r_2$ to $r_1$. 
    The expression $c_1$ represents a symbol cluster, that relays the input from $r_1$ to $r_2$ instead of directly connecting $r_1$ to $r_2$.
    This way, when $c$ sends an inhibition signal to $c_1$ the connection between $r_1$ and $r_2$ is cut, 
    likewise for $c_2$.
    All the arrows in solid black are assignation connections. 
    The arrows with the dotted line directed away from $c$ represent strong inhibition signals that can be sent to $c_1$, $c_2$, 
    $r_1$ or $r_2$ with the effect of clearing the value of $r_1$ or $r_2$ if targetting them or blocking the connection between $r_1$ and $r_2$ through $c_1$ or $r_2$ and $r_1$ through $c_2$. 
    Normally, the inhibition signal is maintained on $c_1$ and $c_2$ so that the value of $r_1$ does not interfere with the value of $r_2$ and vice-versa, 
    whereas no inhibition signal are sent to $r_1$ and $r_2$ with the result that they maintain their value.
    To tranfer a value from $r_1$ to $r_2$, 
    $c$ sends an inhibition signal to $r_2$ to clear its value and then temporily releases the inhibition signal on $c_1$ enabling the content of $r_1$ to flow to $c_1$ and then to $r_2$. 
    Similarly, $c$ may send the content of $r_2$ to $r_1$ by clearing $r_1$ and enabling $c_2$.

    \begin{figure}
        \centering
    
        \begin{tikzpicture}[->,>=stealth,shorten >=1pt,auto,node distance=2.8cm,thick]
     
            \tikzstyle register=[draw, minimum width=1cm, line width=1pt]
            \tikzstyle cluster=[draw, minimum width=1cm, line width=1pt]
            \tikzstyle control=[draw, rounded corners, blue, minimum width=1cm, line width=1pt]
    
            \node[register] (r1)                         {$r_1$};
            \node[red]      (c1)     [below=0.975cm of r1] {};
            \node[register] (r2)     [below=0.975cm of c1] {$r_2$};
            \node[control]  (c2)     [below=1.9cm of r2] {$c$};
            \node[register] (r3)     [below=1.9cm of c2] {$r_3$};
            \node[red]      (c3)     [below=0.975cm of r3] {};
            \node[register] (r4)     [below=0.975cm of c3] {$r_4$};
            \node (r4c)     [below=1cm of r4] {};
            \node (r4l)     [left=5cm of r4c] {};
            \node (r4r)     [right=5cm of r4c] {};
            \node (Wo1234i1234) [below=0cm of r4c] {$W_{o_7,i7}$};
            \node[cluster]  (o1)     [right=2cm of r1] {$o_1$};
            \node[cluster]  (o2)     [right=2cm of r2] {$o_2$};
            \node[cluster]  (o3)     [right=2cm of r3] {$o_3$};
            \node[cluster]  (o4)     [right=2cm of r4] {$o_4$};
            \node[cluster]  (o12)    [right=3.5cm of c1] {$o_5$};
            \node[cluster]  (o34)    [right=3.5cm of c3] {$o_6$};
            \node[cluster]  (o1234)  [right=5.5cm of c2] {$o_7$};
            \node  (o1234c)  [right=1cm of o1234] {};
            \node[cluster]  (i12)    [left=3.5cm of c1] {$i_5$};
            \node[cluster]  (i34)    [left=3.5cm of c3] {$i_6$};
            \node[cluster]  (i1234)  [left=5.5cm of c2] {$i_7$};
            \node  (i1234c)  [left=1cm of i1234] {};
            \node[cluster]  (i1)     [left=2cm of r1] {$i_1$};
            \node[cluster]  (i2)     [left=2cm of r2] {$i_2$};
            \node[cluster]  (i3)     [left=2cm of r3] {$i_3$};
            \node[cluster]  (i4)     [left=2cm of r4] {$i_4$};
             
            \path
                    (r1)    edge [loop above] node {$W_{r_1,r_1}$} (r1)
                    (r2)    edge [loop above] node {$W_{r_2,r_2}$} (r2)
                    (r3)    edge [loop below] node {$W_{r_3,r_3}$} (r3)
                    (r4)    edge [loop below] node {$W_{r_4,r_4}$} (r4)
                    (o1)    edge              node {$W_{o_1,o_5}$} (o12)
                    (o2)    edge              node {$W_{o_2,o_5}$} (o12)
                    (o3)    edge [left]       node {$W_{o_3,o_6}$} (o34)
                    (o4)    edge [right]              node {$W_{o_4,o_6}$} (o34)
                    (o12)   edge              node {$W_{o_5,o_7}$} (o1234)
                    (o34)   edge              node {$W_{o_6,o_7}$} (o1234);
    
            \path   (r1)    edge              node {$W_{r_1,o_1}$} (o1)
                    (r2)    edge[pos=0.22]              node {$W_{r_2,o_2}$} (o2)
                    (r3)    edge[pos=0.7]              node {$W_{r_3,o_3}$} (o3)
                    (r4)    edge              node {$W_{r_4,o_4}$} (o4)
                    (i1)    edge              node {$W_{i_1,r_1}$} (r1)
                    (i2)    edge[pos=0.22]              node {$W_{i_2,r_2}$} (r2)
                    (i3)    edge[pos=0.3]              node {$W_{i_3,r_3}$} (r3)
                    (i4)    edge              node {$W_{i_4,r_4}$} (r4)
                    (i12)   edge              node {$W_{i_5,i_1}$} (i1)
                    (i12)   edge              node {$W_{i_5,i_2}$} (i2)
                    (i34)   edge[right]              node {$W_{i_6,i_3}$} (i3)
                    (i34)   edge[left]              node {$W_{i_6,i_4}$} (i4)
                    (i1234) edge              node {$W_{i_7,i_5}$} (i12)
                    (i1234) edge              node {$W_{i_7,i_6}$} (i34);
    
            \draw  [->,thick] (o1234.east) [rounded corners=10pt] -- (o1234c.east) [rounded corners=10pt] -- (r4r.south) [rounded corners=10pt] -- (r4l.south) [rounded corners=10pt] -- (i1234c.west) [rounded corners=10pt] -- (i1234.west);
                    
    
            \path [red, loosely dotted, line width=2pt]
                    (c2)     edge[bend left, pos=0.8]              node {$W_{c,r_1}$} (r1)
                    (c2)     edge[pos=0.8, right]              node {$W_{c,r_2}$} (r2)
                    (c2)     edge[pos=0.8, left]              node {$W_{c,r_3}$} (r3)
                    (c2)     edge[bend left, pos=0.8, left]              node {$W_{c,r_4}$} (r4)
                    (c2)     edge[pos=0.9]             node {$W_{c,i_1}$} (i1)
                    (c2)     edge              node {$W_{c,i_2}$} (i2)
                    (c2)     edge[left]              node {$W_{c,i_3}$} (i3)
                    (c2)     edge[pos=0.8]              node {$W_{c,i_4}$} (i4)
                    (c2)     edge[pos=0.8]              node {$W_{c,o_1}$} (o1)
                    (c2)     edge[right]              node {$W_{c,o_2}$} (o2)
                    (c2)     edge              node {$W_{c,o_3}$} (o3)
                    (c2)     edge[pos=0.8, left]              node {$W_{c,o_4}$} (o4);

            \node (cluster0)   [above=2.5cm of o1.west]      {};
            \node (cluster1)    [right=2cm of cluster0] {};
            \node (cluster2)    [right=0cm of cluster1] {Symbol Cluster};
            \node[cluster] (cluster3)    [left=0cm of cluster1] {name};
            \node (bcluster0) [below=0.3cm of cluster0] {};
            \node (bcluster1)    [right=2cm of bcluster0] {};
            \node (bcluster2)    [right=0cm of bcluster1] {Panel Cluster};
            \node[control] (bcluster3)    [left=0cm of bcluster1] {name};
            \node (assignation0) [below=0.3cm of bcluster0] {};
            \node (assignation1)    [right=2cm of assignation0] {};
            \node (assignation2)    [right=0cm of assignation1] {Assignation Connection};
            \node (control0) [below=0.3cm of assignation0] {};
            \node (control1)    [right=2cm of control0] {};
            \node (control2)    [right=0cm of control1] {Control Connection};
    
            \path[->,>=stealth,shorten >=1pt,auto,node distance=2.8cm,thick]   
            (assignation0)    edge              node {} (assignation1)
            (control0) edge [red, loosely dotted, line width=2pt]     node {} (control1);

        \end{tikzpicture}         
        \caption{Register switch box with four registers}
        \label{figRegisterSwitchbox}
    \end{figure}
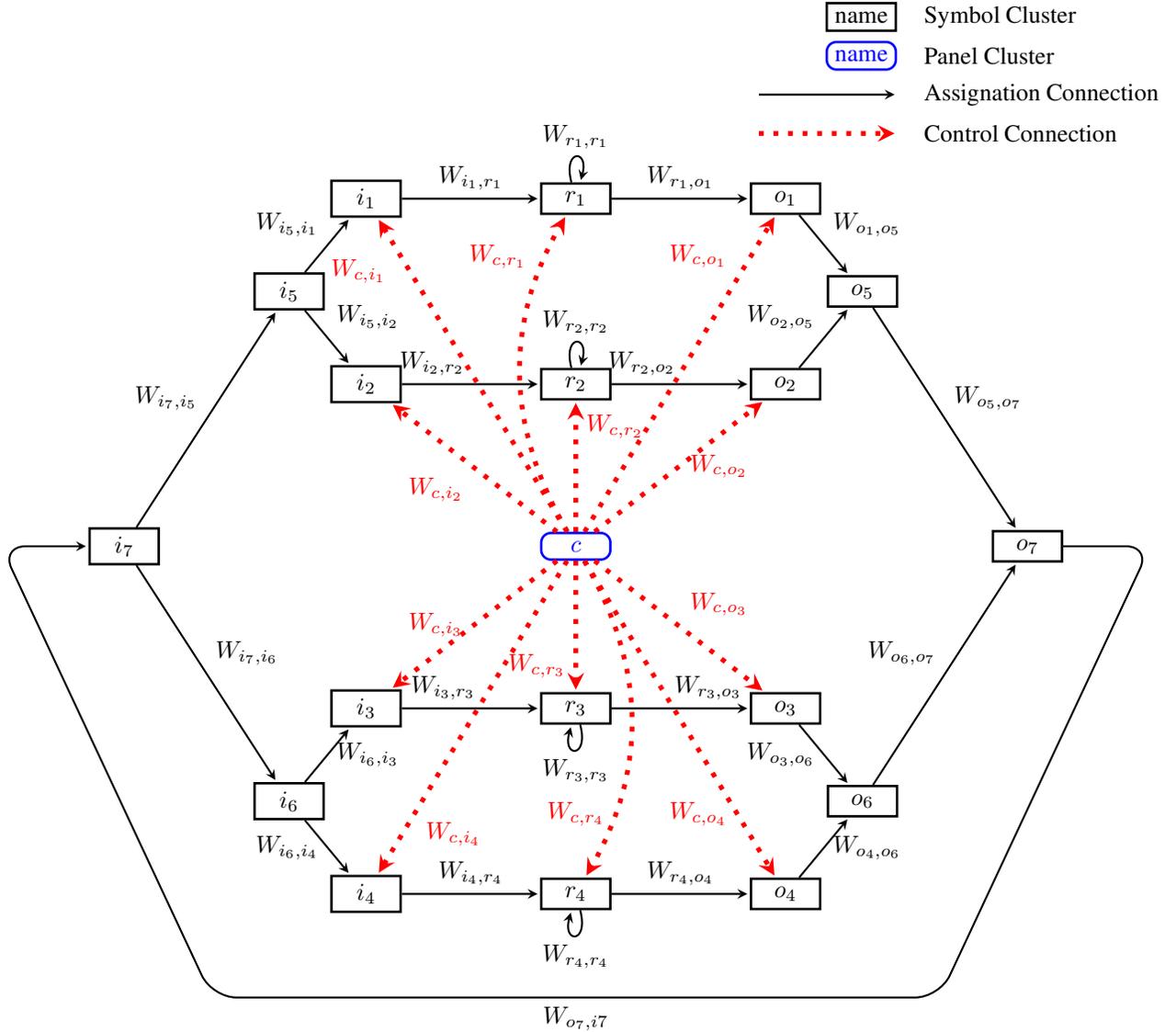

    We extend the circuit to a register switch box containing multiple registers to move the content of one register to another.  
    It is possible to organize the registers in two hierarchies: output one and input one, thereby avoiding a quadratic explosion on the number of connections.
    The register switch box in figure \ref{figRegisterSwitchbox} can move symbols across 4 registers while minimizing the number of connections. 
    In this case, normally an inhibition signal is sent to $i_1$, $i_2$, $i_3$, $i_4$, 
    and $o_1$, $o_2$, $o_3$, $o_4$ so that the values of $r_1$, $r_2$, $r_3$ and $r_4$ do not interfere with each other. 
    Let us say we want to transfert a value from $r_1$ to $r_4$. 
    First, an inhibition signal is sent to $r_4$ to clear its value preparing it to receive a new value. 
    Then the inhibition signal on $o_1$ and $i_4$ are temporarily released allowing the value of $r_1$ to flow to $r_4$. 
    And that's it, of course the network of figure \ref{figRegisterSwitchbox} may be extended to support more registers.

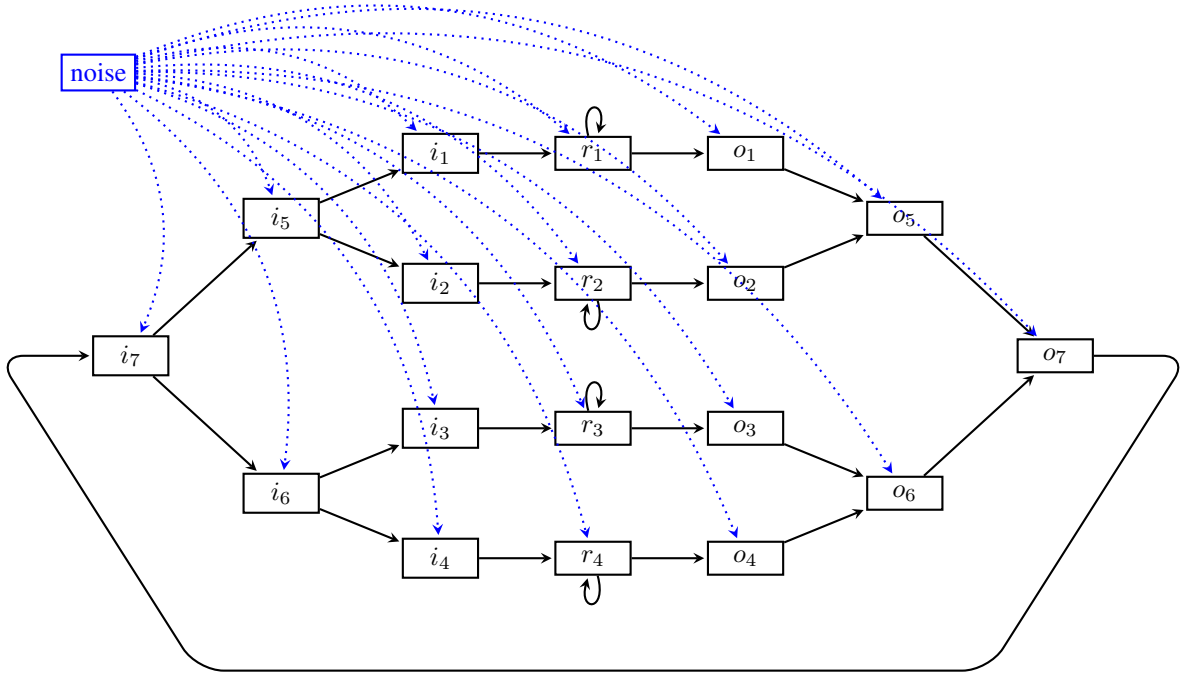
\begin{figure}
        \centering
        \begin{tikzpicture}[->,>=stealth,shorten >=1pt,auto,node distance=2.8cm,thick]
     
            \tikzstyle register=[draw, minimum width=1cm]
            \tikzstyle control=[draw, minimum width=1cm]
    
            \node[register] (r1)                         {$r_1$};
            \node[red]      (c1)     [below=0.5cm of r1] {};
            \node[register] (r2)     [below=0.5cm of c1] {$r_2$};
            \node           (c2)     [below=0.6cm of r2] {};
            \node[register] (r3)     [below=0.6cm of c2] {$r_3$};
            \node[red]      (c3)     [below=0.5cm of r3] {};
            \node[register] (r4)     [below=0.5cm of c3] {$r_4$};
            \node[control]  (o1)     [right=1cm of r1] {$o_1$};
            \node[control]  (o2)     [right=1cm of r2] {$o_2$};
            \node[control]  (o3)     [right=1cm of r3] {$o_3$};
            \node[control]  (o4)     [right=1cm of r4] {$o_4$};
            \node[control]  (o12)    [right=3.5cm of c1] {$o_5$};
            \node[control]  (o34)    [right=3.5cm of c3] {$o_6$};
            \node[control]  (o1234)  [right=5.5cm of c2] {$o_7$};
            \node[control]  (i12)    [left=3.5cm of c1] {$i_5$};
            \node[control]  (i34)    [left=3.5cm of c3] {$i_6$};
            \node[control]  (i1234)  [left=5.5cm of c2] {$i_7$};
            \node[control]  (i1)     [left=1cm of r1] {$i_1$};
            \node[control]  (i2)     [left=1cm of r2] {$i_2$};
            \node[control]  (i3)     [left=1cm of r3] {$i_3$};
            \node[control]  (i4)     [left=1cm of r4] {$i_4$};
            \node[blue, draw]     (seed)   [above left=2cm of i12] {noise};

            \path
                    (r1)    edge [loop above] node {} (r1)
                    (r2)    edge [loop below] node {} (r2)
                    (r3)    edge [loop above] node {} (r3)
                    (r4)    edge [loop below] node {} (r4)
                    (o2)    edge              node {} (o12)
                    (o4)    edge              node {} (o34)
                    (o34)   edge              node {} (o1234);
    
            \path   (r1)    edge              node {} (o1)
                    (r2)    edge              node {} (o2)
                    (r3)    edge              node {} (o3)
                    (r4)    edge              node {} (o4)
                    (i1)    edge              node {} (r1)
                    (i2)    edge              node {} (r2)
                    (i3)    edge              node {} (r3)
                    (i4)    edge              node {} (r4)
                    (i12)   edge              node {} (i1)
                    (i12)   edge              node {} (i2)
                    (i34)   edge              node {} (i3)
                    (i34)   edge              node {} (i4)
                    (i1234) edge              node {} (i12)
                    (i1234) edge              node {} (i34)
                    (o1)    edge              node {} (o12)
                    (o3)    edge              node {} (o34)
                    (o12)   edge              node {} (o1234);
                    
            \path [blue, dotted]
            (seed)  edge[bend left]              node {} (r1)
            (seed)  edge[bend left]              node {} (r2)
            (seed)  edge[bend left]              node {} (r3)
            (seed)  edge[bend left]              node {} (r4)
            (seed)  edge[bend left]              node {} (i1)
            (seed)  edge[bend left]              node {} (i2)
            (seed)  edge[bend left]              node {} (i3)
            (seed)  edge[bend left]              node {} (i4)
            (seed)  edge[bend left]              node {} (i12)
            (seed)  edge[bend left]              node {} (i34)
            (seed)  edge[bend left]              node {} (i1234)
            (seed)  edge[bend left]              node {} (o1)
            (seed)  edge[bend left]              node {} (o2)
            (seed)  edge[bend left]              node {} (o3)
            (seed)  edge[bend left]              node {} (o4)
            (seed)  edge[bend left]              node {} (o12)
            (seed)  edge[bend left]              node {} (o34)
            (seed)  edge[bend left]              node {} (o1234);
    
            \node (r4c)     [below=1cm of r4] {};
            \node (r4l)     [left=5cm of r4c] {};
            \node (r4r)     [right=5cm of r4c] {};
            \node  (o1234c)  [right=1cm of o1234] {};
            \node  (i1234c)  [left=1cm of i1234] {};
            \draw  [->,thick] (o1234.east) [rounded corners=10pt] -- (o1234c.east) [rounded corners=10pt] -- (r4r.south) [rounded corners=10pt] -- (r4l.south) [rounded corners=10pt] -- (i1234c.west) [rounded corners=10pt] -- (i1234.west);
    
        \end{tikzpicture}         
        \caption{Register switch box receiving noise for learning symbols}
        \label{figLearnRegisterSwitchbox}
    \end{figure}

    Assuming prime attractors are a reality for the brain, we take advantage that prime attractors are random noise to train the assignation connections.
    In figure \ref{figLearnRegisterSwitchbox}, imagine that a source of noise projects parts of its image on all the symbol clusters of the register switch box in such a way that each symbol cluster receives a different part of the noise image.
    Then have the assignation connections (and the self-connections) on the register switch box learn the mapping from the part of the image projected on its input cluster to the part of the image projected on its output cluster.
    Repeat this process multiple times with the source of noise projecting a different noise image.
    What is essential to notice is that the part of the image projected on any symbol cluster uniquely identifies the image.
    Also, given that a cluster of the register switch box holds its projection of the part of a given noise image, an outgoing assignation connection would assign to its output cluster its part of the same noise image.
    Each noise image represents a symbol transferrable from register to register through the register switch box.

    Now that we can transfer symbols between registers, which was the last component needed, we are ready to implement our symbolic computer, which uses a register switch box at its core.
    
\section{Building a symbolic computer} \label{buildingacomputer}

    This section shows how to build a symbolic computer out of spiking neurons using our symbolic framework and components.
    The presumption is that a machine that can execute a program written in the assembly language introduced here is indeed a symbolic computer.
    The explanation of the architecture that interprets it follows with a discussion on the initialization of the symbolic computer. 
    How to program it and the results of tests we made conclude this section. 

\subsection{Assembly langage} \label{AssemblyLanguage}

    Our symbolic computer executes a program defined by sequences of instructions by performing one at the time the action prescribed by each instruction.
    In general, an instruction performs its action by acting on the registers of our symbolic computer.
    
    The symbolic computer maintains 16 registers which take values from the same symbol space.
    Each register has a different function and a name:
    \begin{description}[align=left,labelwidth=2cm]
        \item [reserved] is a special purpose register used implicitly by some instructions;
        \item [r1] is a general purpose register;
        \item [r2] is a general purpose register;
        \item [table] holds the current hash table;
        \item [key] holds the current key for the current hash table;
        \item [value] holds the current value of the current hash table at the current key; 
        \item [code] holds the next instruction to execute;
        \item [code2] holds the default next instruction to execute bound to the executing instruction;
        \item [cont] holds the continuation instruction, that is, the intruction to execute once the current function returns;
        \item [arg] holds an optional value set by the executing instruction;
        \item [alloc] holds the head of the linked list of free symbols;
        \item [alloc2] holds the item link to the head of the linked list of free symbols;
        \item [cons] holds the current cons cell\footnote{The name \emph{cons}, \emph{car} and \emph{cdr} refer to the corresponding lisp functions. Essentially, a \emph{cons} cell is a structure with two values, the \emph{car} value and the \emph{cons} value.  It is used to implement linked list and other data structures. We borrow the concept here. In our case the \emph{cons} cell, and its \emph{car} and \emph{cdr} values are represented by symbols. };
        \item [car] holds the car value of the current cons cell;
        \item [cdr] holds the cdr value of the current cons cell;
        \item [stack] holds the head of the stack that is usually built with the \emph{cons} register.
    \end{description}

    Apart from these 16 registers, the symbolic computer has an input channel of symbols and an output channel.
    It also maintains a list of free symbols and has a special symbol, the \emph{false} symbol.

    An instruction is a symbol on which is bound: 
    \begin{itemize}
        \item an opcode to execute if the \emph{value} register does not contain the \emph{false} symbol,
        \item an opcode to execute otherwise,
        \item an optional symbol to store in the \emph{arg} register,
        \item the next instruction to execute to store in the \emph{code2} register.
    \end{itemize}
    The representation of a program inside the machine is a graph of such instructions. 
    Note that it is a graph and not simply a linked list because the optional value destined to the \emph{arg} register might be an instruction as well as the next instruction to be stored in the \emph{code2} register.
    
    When an instruction held by the \emph{code} register executes, it sets the \emph{code2} register and optionally the \emph{arg} register.
    Then it chooses which opcode to execute depending on the value of the \emph{value} register, moves the content of the \emph{code2} register in the \emph{code} register, and executes the opcode.
    The execution of the opcode might change the value of the \emph{code} register, typically by moving the content of the \emph{arg} register or \emph{cont} register to it.
    In any case, the cycle repeats unless the opcode is the \emph{exit} opcode, in which case the program terminates.

    The following opcodes represents the actions our symbolic computer can do:
    \begin{description}[align=left,labelwidth=2cm]
        \item [alloc-recall] given that the \emph{alloc} register is the head of the linked list of free symbols, sets in the \emph{alloc2} register the next symbol after the head; 
        \item [alloc-bind] given that the \emph{alloc2} register is the head of the linked list of free symbols, sets the \emph{alloc} register as the new head;
        \item [alloc-unbind] given that the \emph{alloc} register is the head of the linked list of free symbols and the \emph{alloc2} register the next symbol after the head, removes the value of the \emph{alloc} register from the list of free symbols, which sets the content of the \emph{alloc2} register as the new head of the list of free symbols;
        \item [cons-recall] given that the \emph{cons} register holds a symbol to which a \emph{car} value and a \emph{cdr} value have been bound to it, sets the \emph{car} register with the \emph{car} value and the \emph{cdr} register with the \emph{cdr} value;
        \item [cons-bind] given that the \emph{cons} register holds a symbol to which a \emph{car} value and a \emph{cdr} value have not been bound to it, binds the value of the \emph{car} register as its \emph{car} value and the value of the \emph{cdr} register as its \emph{cdr} value;
        \item [cons-unbind] given that the \emph{cons} register holds a symbol to which the value of \emph{car} register is bound to its \emph{car} value and the value of its \emph{cdr} register is bound to its \emph{cdr} value, unbinds them;
        \item [table-recall] given that the pair of values of the \emph{table} and \emph{key} registers has a value bound to it, sets the \emph{value} register with that value;
        \item [table-bind] given that the pair of values of the \emph{table} and \emph{key} registers does not have a value bound to it, binds the value of the \emph{value} register to it;
        \item [table-unbind] given that the pair of values of the \emph{table} and \emph{key} registers has the value of the \emph{value} register bound to it, unbinds it;
        \item [read] reads the next value of the input channel into the \emph{reserved} register;
        \item [write] prints the value of the \emph{reserved} register in the output channel;
        \item [nop] does nothing;
        \item [exit] exits the program;
        \item [mov($A$, $B$)] replaces the content of register $B$ by the content of register $A$.
    \end{description}

    The description of the assembly language is now complete. 
    The architecture of the symbolic machine that can execute it follows.

\subsection{Architecture} \label{Architecture}

\begin{figure}
    \centering
    \begin{tikzpicture}[]

        \tikzstyle cluster=[draw, minimum width=1.45cm, line width=1.0pt]
        \tikzstyle bcluster=[draw, minimum width=1.45cm, rounded corners, blue, line width=1.0pt]
        \tikzstyle control=[draw, circle, minimum size=1cm]

        \node[cluster] (reserved)                          {reserved};
        \node[cluster] (stack)   [below=0.4cm of reserved] {stack};
        \node[cluster] (cont)    [below=0.4cm of stack]    {cont};
        \node[cluster] (r1)      [below=0.4cm of cont]     {$r_1$};
        \node[cluster] (r2)      [below=0.4cm of r1]       {$r_2$};
        \node[cluster] (table)   [below=0.4cm of r2]       {table};
        \node[cluster] (key)     [below=0.4cm of table]    {key};
        \node[cluster] (value)   [below=0.4cm of key]      {value};
        \node[cluster] (code)    [below=0.4cm of value]    {code};
        \node[cluster] (coden)   [below=0.4cm of code]     {code2};
        \node[cluster] (arg)     [below=0.4cm of coden]    {arg};
        \node[cluster] (alloc)   [below=0.4cm of arg]      {alloc};
        \node[cluster] (allocn)  [below=0.4cm of alloc]    {alloc2};
        \node[cluster] (cons)    [below=0.4cm of allocn]   {cons};
        \node[cluster] (car)     [below=0.4cm of cons]     {car};
        \node[cluster] (cdr)     [below=0.4cm of car]      {cdr};
        \node (swanchor) [above=0.0cm of reserved] {};
        \node[draw, rotate=90, minimum width=15.2cm, minimum height=3cm, red, line width=2pt]           (sw)      [left=2.1cm of swanchor]       {register switch box};
        \node[cluster] (consc)   [right=2cm of cons]       {cons-c}; 
        \node[cluster] (codec)   [right=2cm of code]       {code-c}; 
        \node[cluster] (allocc)  [right=2cm of alloc]      {alloc-c}; 
        \node[cluster] (hash)    [right=4cm of table]      {hash};
        \node (sdn) [above=0.55cm of hash] {\textbf{H}};
        \node[bcluster] (test)    [right=2cm of value]      {test};
        \node[cluster] (opcodeeq)[right=1cm of codec]      {opcode-eq};
        \node[cluster] (opcodeneq)[below=0.4cm of opcodeeq]   {opcode-neq};
        \node[cluster] (opcode)  [right=1cm of opcodeeq]   {opcode};
        \node[bcluster] (microinstruction) [right=6cm of alloc] {microinstruction};
        \node[bcluster, minimum width=2.45cm] (controlc) [below=1.8cm of microinstruction] {control};
 
        \path[->,>=stealth,shorten >=1pt,auto,node distance=2.8cm,thick]   
                (cons)    edge [loop above] node {} (cons)
                (car)     edge [loop above] node {} (car)
                (cdr)     edge [loop above] node {} (cdr)
                (stack)   edge [loop above] node {} (stack)
                (code)    edge [loop above] node {} (code)
                (coden)   edge [loop above] node {} (coden)
                (alloc)   edge [loop above] node {} (alloc)
                (allocn)  edge [loop above] node {} (allocn)
                (table)   edge [loop above] node {} (table)
                (key)     edge [loop above] node {} (key)
                (value)   edge [loop above] node {} (value)
                (reserved)edge [loop above] node {} (reserved)
                (arg)     edge [loop above] node {} (arg)
                (r1)      edge [loop above] node {} (r1)
                (r2)      edge [loop above] node {} (r2)
                (cont)    edge [loop above] node {} (cont)
                (test)    edge [loop above, blue, densely dotted] node {} (test)
                (microinstruction) edge [loop right, blue, densely dotted] node {} (microinstruction)
                (cons)    edge              node {} (consc)
                (consc)   edge [dash pattern=on 4pt off 4pt, black!60!green, line width=1.25pt]              node {} (car)
                (consc)   edge [dash pattern=on 4pt off 4pt, black!60!green, line width=1.25pt]             node {} (cdr)
                (code)    edge              node {} (codec)
                (codec)   edge [dash pattern=on 8pt off 2pt on 2pt off 2pt]              node {} (coden)
                (codec)   edge [dash pattern=on 8pt off 2pt on 2pt off 2pt]              node {} (arg)
                (alloc)   edge              node {} (allocc)
                (allocc)  edge [dash pattern=on 4pt off 4pt, black!60!green, line width=1.25pt]             node {} (allocn)
                (table)   edge              node {} (sdn)
                (key)     edge              node {} (sdn)
                (sdn)     edge [snake=coil, line after snake=4pt]             node {} (hash)
                (value)   edge [snake=triangles, segment length=5pt, blue]              node {} (test)
                (codec)   edge [dash pattern=on 8pt off 2pt on 2pt off 2pt]              node {} (opcodeeq)
                (codec)   edge [dash pattern=on 8pt off 2pt on 2pt off 2pt]              node {} (opcodeneq)
                (opcodeeq)edge              node {} (opcode)
                (opcodeneq) edge            node {} (opcode)
                (opcode)  edge [snake=triangles, segment length=5pt, blue]      node {} (microinstruction)
                (microinstruction) edge [blue, densely dotted]     node {} (controlc);

        \path [->,>=stealth,shorten >=1pt,auto,node distance=2.8cm, red, loosely dotted, line width=2pt]
                (test)    edge              node {} (opcodeeq)
                (test)    edge              node {} (opcodeneq)
                (controlc) edge              node {} (codec)
                (controlc) edge              node {} (consc)
                (controlc) edge              node {} (allocc);

        \draw [->,thick,dash pattern=on 4pt off 4pt, black!60!green, line width=1.25pt] (hash.west) to (value.north east);
        \draw [->,red, loosely dotted, line width=2pt] (controlc.west) to [out=200,in=300] (sw.west);
        \draw [->,red, loosely dotted, line width=2pt] (controlc.east) to [out=0,in=0] (hash.east);


        \node (cluster0anchor) [above=5.5cm of hash]      {};
        \node (cluster0)   [right=0cm of cluster0anchor]      {};
        \node (cluster1)    [right=2cm of cluster0] {};
        \node (cluster2)    [right=0cm of cluster1] {Symbol Cluster};
        \node[cluster] (cluster3)    [left=0cm of cluster1] {name};
        \node (bcluster0) [below=0.3cm of cluster0] {};
        \node (bcluster1)    [right=2cm of bcluster0] {};
        \node (bcluster2)    [right=0cm of bcluster1] {Panel Cluster};
        \node[bcluster] (bcluster3)    [left=0cm of bcluster1] {name};
        \node (assignation0) [below=0.3cm of bcluster0] {};
        \node (assignation1)    [right=2cm of assignation0] {};
        \node (assignation2)    [right=0cm of assignation1] {Assignation Connection};
        \node (mapping0) [below=0.3cm of assignation0] {};
        \node (mapping1)    [right=2cm of mapping0] {};
        \node (mapping2)    [right=0cm of mapping1] {Mapping Connection};
        \node (binding0) [below=0.3cm of mapping0] {};
        \node (binding1)    [right=2cm of binding0] {};
        \node (binding2)    [right=0cm of binding1] {One-shot Symbolic Memory};
        \node (secdegree0) [below=0.3cm of binding0] {};
        \node (secdegree1)    [right=2cm of secdegree0] {};
        \node (secdegree2)    [right=0cm of secdegree1] {Second-degree Connection};
        \node (secdegree0up) [above=0cm of secdegree0] {};
        \node (secdegree0down) [below=0cm of secdegree0] {};
        \node (secdegreeH) [right=1cm of secdegree0] {\textbf{H}};
        \node (control0) [below=0.3cm of secdegree0] {};
        \node (control1)    [right=2cm of control0] {};
        \node (control2)    [right=0cm of control1] {Control Connection};
        \node (bitwise0) [below=0.3cm of control0] {};
        \node (bitwise1)    [right=2cm of bitwise0] {};
        \node (bitwise2)    [right=0cm of bitwise1] {Per-neuron Connection};
        \node (recognizer0) [below=0.3cm of bitwise0] {};
        \node (recognizer1)    [right=2cm of recognizer0] {};
        \node (recognizer2)    [right=0cm of recognizer1] {Recognizer Connection};

        \path[->,>=stealth,shorten >=1pt,auto,node distance=2.8cm,thick]   
        (assignation0)    edge              node {} (assignation1)
        (binding0)   edge [dash pattern=on 4pt off 4pt, black!60!green, line width=1.25pt]              node {} (binding1)
        (mapping0)   edge [dash pattern=on 8pt off 2pt on 2pt off 2pt]              node {} (mapping1)
        (secdegreeH)     edge [snake=coil, line after snake=4pt]             node {} (secdegree1)
        (secdegree0up)    edge              node {} (secdegreeH)
        (secdegree0down)    edge              node {} (secdegreeH)
        (recognizer0)   edge [snake=triangles, segment length=5pt, blue]              node {} (recognizer1)
        (bitwise0) edge [blue, densely dotted]     node {} (bitwise1)
        (control0) edge [red, loosely dotted, line width=2pt]     node {} (control1);

    \end{tikzpicture}         
    \caption{Architecture of our neuro-symbolic computer}
    \label{figArchitecture}
\end{figure}
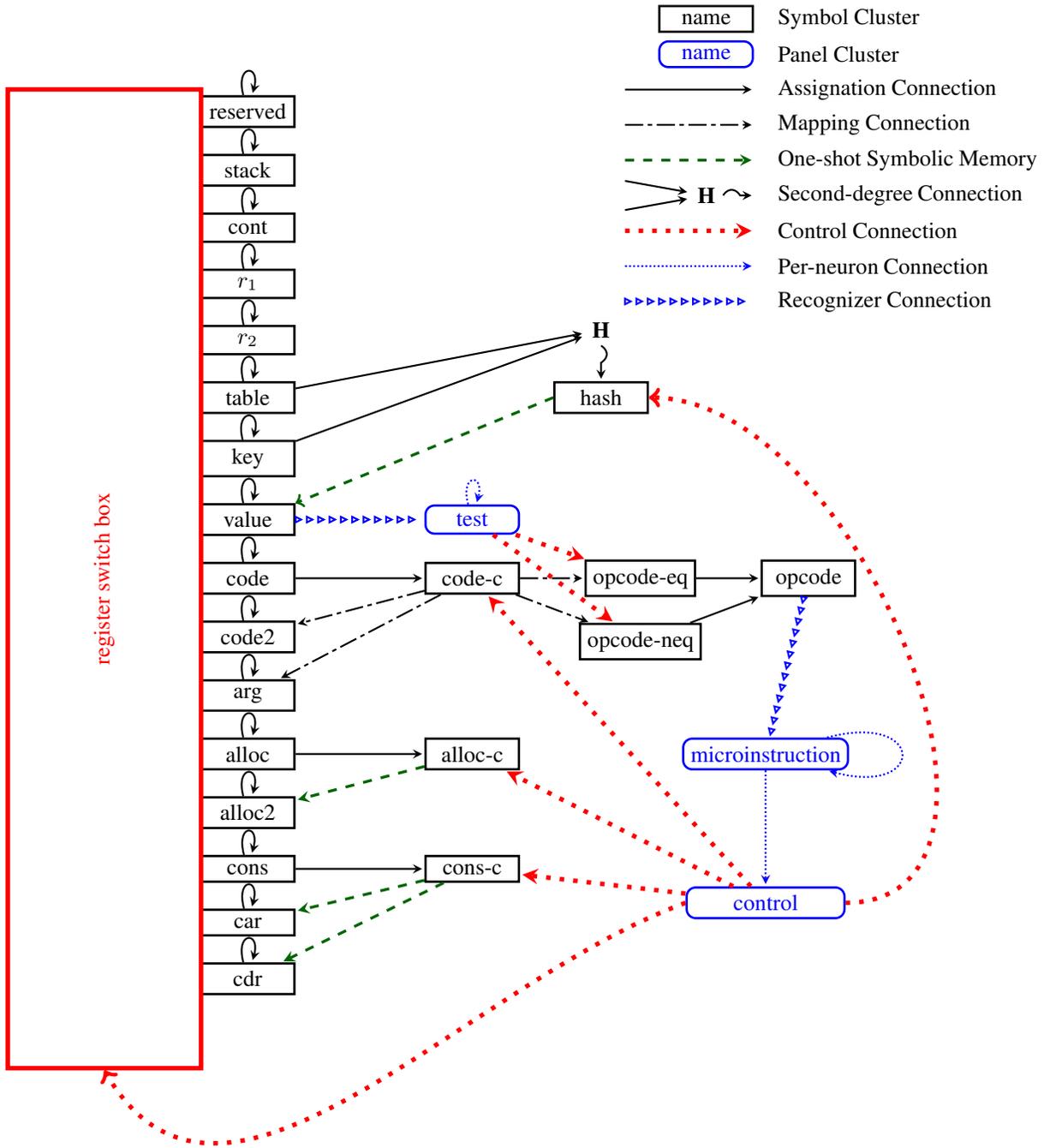
    Figure \ref{figArchitecture} shows the architecture. 
    
    The main component on the left is a register switch box with the 16 registers mentioned earlier. 
    
    There is also a \emph{microinstruction} cluster that runs sequences of microinstructions.
    Each sets the \emph{control} cluster to control the flow of inhibition signals and binding and unbinding signals (dotted lines) entering the various parts.  
    For instance, 
    inhibition signals enter the register switch box either to control the transfer of symbols between registers or to clear a 
    register.  
    For simplicity, a single neuron represents each microinstruction in the \emph{microinstruction} cluster. 
    It triggers the next microinstruction, thereby creating a sequence of microinstructions executing in a row\footnote{
        There is an exception to this with the first neuron, which also activates a barrier neuron that will prevent reentry by inhibiting the second neuron after the first time}. 
    A sequence of neurons represents each sequence of microinstructions, each microinstruction prompting the next one. 
    The neuron associated with a microinstruction, 
    when activated, fires the neurons in the \emph{control} cluster. 
    These neurons, in turn, send the inhibition, binding, and unbinding signals to perform the function of the microinstruction.
    Two special microinstructions enable communication between the simulated symbolic computer and the actual computer:
    a \emph{read} microinstruction to read a value into a register and a \emph{write} microinstruction to print the value of a register.

    Finally, some registers have a special purpose circuitry associated with them:

    \begin{itemize}
        \item An assignation connection links the \emph{alloc} register to the \emph{alloc-c} cluster whereas a one-shot symbolic memory connection links the \emph{alloc-c} cluster to the \emph{alloc2} register.
        However, an inhibition signal sent by the \emph{microinstruction} cluster disables the \emph{alloc-c} cluster the majority of the time.  
        During a recall, the \emph{microinstruction} cluster first clear the
        \emph{alloc2} register with an inhibition signal through the register switch box. 
        It then enables the \emph{alloc-c} cluster by temporarily blocking the inhibition signal so that the \emph{alloc-c} cluster can set the \emph{alloc2} register. 
        Instead of inhibiting the \emph{alloc-c} cluster, the \emph{microinstruction cluster} can also send it a binding message, thereby binding the symbol held by the \emph{alloc2} register to the \emph{alloc-c} cluster.
        Similarly, the \emph{microinstruction cluster} can send an unbinding message to the \emph{alloc-c} cluster instead of inhibiting it, thereby unbinding the symbol held by the \emph{alloc2} register from the \emph{alloc-c} cluster.
        
        \item The same mechanism is used to recall/bind/unbind the \emph{car} register and the \emph{cdr} register to the \emph{cons} register.

        \item The \emph{table} register and the \emph{key} register
        feed a second-degree random network that outputs its result to the \emph{hash} cluster. 
        However, an inhibition signal sent by the \emph{microinstruction} cluster disables the \emph{hash} cluster most of the time.  
        During a recall, the \emph{microinstruction} cluster first clears the \emph{value} register with an inhibition signal through the register switch box. 
        It then enables the \emph{hash} cluster by temporarily blocking the inhibition signal so that the \emph{hash} cluster can set the \emph{value} register. 
        The \emph{microinstruction cluster} can also trigger the binding of the symbol held by the \emph{table} register in pair with the symbol held by the \emph{key} register to the symbol held by the \emph{value} register.
        This is accomplished by sending the \emph{hash} cluster a binding signal instead of an inhibition signal.
        Similarly the \emph{microinstruction cluster} can trigger the unbinding of the \emph{value} by sending the \emph{hash} cluster an unbinding signal.
        The hash table also supports a default value; that is, there is a second connection going from the \emph{hash} cluster to the \emph{value} register.
        This second connection feeds the default prime attractor, the \emph{false} symbol, with half intensity, with the result that the \emph{false} symbol is selected when no value is bound to the \emph{table} \emph{key} pair.

        \item The \emph{value} register has special circuitry to test when its content is the \emph{false} symbol. 
        The connection fires a given neuron of the \emph{test} cluster, the \emph{false} neuron, when the neurons of the \emph{false} value are excited.
        The \emph{test} cluster has a feedback connection to fire another neuron, the \emph{true} neuron, when the \emph{false} neuron is not firing.
        We use the firing of the \emph{true} and \emph{false} neurons to control the flow of execution, as we see shortly.
        For simplicity, we use a single \emph{true} and \emph{false} neuron.
    
        \item The \emph{code} register also has special circuitry associated with it. Its value represents the instruction to execute.
        From it we can obtain:
        \begin{itemize}
            \item the next instruction assigned to the \emph{code2} register, 
            \item an arbitrary symbol assigned to the \emph{arg} register, 
            \item and two opcode values assigned to the \emph{opcode-eq} and \emph{opcode-neq} clusters with one inhibited by the \emph{test} cluster and the other flowing into the \emph{opcode} cluster.
        \end{itemize} 
        The \emph{opcode} cluster determines the next microinstructions to execute.
        It triggers a sequence of neurons into the \emph{microinstruction} cluster that begins by inhibiting the \emph{code-c} cluster to clear the \emph{opcode} cluster to avoid reentry; 
        while performing the opcode function going through the sequence. 
        The sequence terminates by setting up the execution of the following upcode. 
        To do so, the sequence clears the \emph{code2} and \emph{arg} registers. 
        Then, the sequence temporarily block the inhibition signal on the \emph{code-c} cluster so that the value of the \emph{code} register can flow into it.
        In turn, 
        the value of \emph{code-c} register will recall new values for:
        \begin{itemize}
            \item the \emph{code2} register, 
            \item the \emph{arg} register,
            \item the \emph{opcode-eq} and \emph{opcode-neq} clusters with one of them flowing to the \emph{opcode} register depending on the \emph{true} and \emph{false} neurons of the \emph{test} cluster. 
        \end{itemize}
        The content of the \emph{opcode} cluster then triggers its sequence of microinstructions.

    \end{itemize}

    It is reasonably straightforward to encode in the sequences of microinstructions the proper activation of the neurons of the \emph{control} cluster to act for each opcode.
    That is:
    \begin{description}[align=left,labelwidth=2cm]
        \item [alloc-recall] recalls the value of the \emph{alloc2} register from the value of the \emph{alloc} register through the \emph{alloc-c} cluster;
        \item [alloc-bind] binds the value of the \emph{alloc2} register to the value of the \emph{alloc} register through the \emph{alloc-c} cluster;
        \item [alloc-unbind] unbinds the value of the \emph{alloc2} register from the value of the \emph{alloc} register through the \emph{alloc-c} cluster;
        \item [cons-recall] recalls the values of the \emph{car} and \emph{cdr} registers from the value of the \emph{cons} register through the \emph{cons-c} cluster;
        \item [cons-bind] binds the values of the \emph{car} and \emph{cdr} registers to the value of the \emph{cons} register through the \emph{cons-c} cluster;
        \item [cons-unbind] unbinds the values of the \emph{car} and \emph{cdr} registers from the value of the \emph{cons} register through the \emph{cons-c} cluster;
        \item [table-recall] recalls the value of \emph{value} register from the pair of values of the \emph{table} and \emph{key} registers through the \emph{hash} cluster;
        \item [table-bind] binds the value of \emph{value} register to the pair of values of the \emph{table} and \emph{key} registers through the \emph{hash} cluster;
        \item [table-unbind] unbinds the value of \emph{value} register from the pair of values of the \emph{table} and \emph{key} registers through the \emph{hash} cluster;
        \item [nop] does nothing;
        \item [mov($A$, $B$)] replaces the content of register $B$ by the content of register $A$.
    \end{description}
    We have implemented this architecture in an experiment that tests this symbolic computer.

    Note that the \emph{read}, \emph{write} and \emph{exit} opcodes are exceptions and are not implemented using the \emph{control} cluster.
    Rather they serve the interaction with the encompassing system.

    For simplicity, the code uses prime attractors to save its content.
    However, since they are a limited resource, one might consider alternatives to restrict their use to entry points.

    All the registers in the register switch box are in the same symbol space in the sense that all the registers can hold a symbol assigned to any of them.
    The \emph{opcode-eq} and \emph{opcode-neq} clusters and the \emph{opcode} register use a different symbol space than the registers of the register switch box.
    That is, a register of the register switch box cannot take the symbol of the \emph{opcode} register as value and vice-versa.
    It is possible to use a distinct switch box for the \emph{code}, \emph{code2} and \emph{cont} registers to have use their own symbol space,
    thereby doubling the capacity. 
    We did not use this architecture for simplicity as it would have meant duplicating \emph{cons}, \emph{car}, \emph{cdr} and \emph{stack} registers and the mechanism associated with them as well.

    Another improvement that could be done is regarding the hash table mechanism that is used to implement global tables like \emph{is-digit}, etc..., the issue is that the same structure is also used in some programs to encode live data, which erodes the hash table content with deletions.
    To remedy the situation, the programs should refresh periodically the content of global tables like the \emph{is-digit} table, etc...; 
    another solution would be to duplicate the \emph{hash} cluster to implement a hash table for global data and a hash table for live data.

    This architecture can execute some simple symbolic programs. Let us see how it is done.

\subsection{Common sequence of instructions} \label{CommonSequences}

    Some sequences of instructions are more frequent and illustrate how to do programming.

    To recapitulate: an instruction is composed of an opcode to execute if the content of the \emph{value} register is not the \emph{false} symbol,
    an opcode to execute otherwise, an optional value for the \emph{arg} register, and the next instruction.
    To simplify the notation, we omit the next instruction entry as we present the instructions in sequence. 
    We give one opcode if the two opcodes are identical, and we do not specify any value for the \emph{arg} register if the opcodes are not using the \emph{arg} register.
    If the program refers to an instruction, we precede the instruction with a label followed by a colon to refer to it.
    If all the values are present will use the notation \textbf{label: (opcode1, opcode2, arg = \emph{argvalue})} for an instruction.

    \begin{description}[align=left,labelwidth=2cm]
        \item [Allocate a symbol into register $r$]:
            \begin{itemize}
                \item perform an \textbf{alloc-recall} to get in the \emph{alloc2} the next symbol;
                \item perform an \textbf{alloc-unbind} to break the link;
                \item perform a \textbf{mov(\emph{alloc}, $r$)} to set $r$ with the new symbol;
                \item perform a \textbf{mov(\emph{alloc2}, \emph{alloc})} to have the register \emph{alloc} point to the next free symbol.
            \end{itemize}
        \item [Deallocate the symbol from register $r$]:
            \begin{itemize}
                \item perform a \textbf{mov(\emph{alloc}, \emph{alloc2})} to have the register \emph{alloc2} point to the rest of free symbol linked-list;
                \item perform a \textbf{mov($r$, \emph{alloc})} to have the content of $r$ be the head of the free symbol linked-list;
                \item perform an \textbf{alloc-bind} to create the link.
            \end{itemize}
        \item [Push the content of register $r$ into the stack]:
            \begin{itemize}
                \item allocate a symbol into the \emph{cons} register;
                \item perform a \textbf{mov($r$, \emph{car})} to have the content of $r$ be first element of the pair;
                \item perform a \textbf{mov(\emph{stack}, \emph{cdr})} to put the stack pointer into the second element of the pair;
                \item perform an \textbf{cons-bind} to bind the cons cell;
                \item perform a \textbf{mov(\emph{cons}, \emph{stack})} to set the new head of the stack.
            \end{itemize}
        \item [Pop the content of the stack into the register $r$]:
            \begin{itemize}
                \item perform a \textbf{mov(\emph{stack}, \emph{cons})} to prepare to unpack the stack head;
                \item perform an \textbf{cons-recall} to unpack the cons cell;
                \item perform an \textbf{cons-unbind} to release the cons cell;
                \item perform a \textbf{mov(\emph{car}, $r$)} to set $r$ with the first element of the pair;
                \item perform a \textbf{mov(\emph{cdr}, \emph{stack})} to put the stack pointer to the next value;
                \item deallocate the symbol from the \emph{cons} register.
            \end{itemize}
        \item [Call the subroutine at instruction $l$]:
            \begin{itemize}
                \item push the content of the \emph{cont} register into the stack;
                \item perform a \textbf{(mov(\emph{arg}, \emph{cont}), \emph{arg} = \emph{return instruction})}, to set the register \emph{cont} with the return instruction;
                \item perform a \textbf{(mov(\emph{arg}, \emph{code}), \emph{arg} = $l$)}, to jump to the subroutine at instruction $l$;
                \item \textbf{\emph{return instruction}}: pop the content of the stack into the \emph{cont} register.
            \end{itemize}

        \item [Return from a subroutine]:
            \begin{itemize}
                \item perform a \textbf{mov(\emph{cont}, \emph{code})} to return from the subroutine.
            \end{itemize}
        
        \item [Do \emph{action} if the \emph{value} register is not the \emph{false} symbol]:
        \begin{itemize}
            \item perform a \textbf{(nop, mov(\emph{arg}, \emph{code}), arg = \emph{branch instruction})}, to branch to \emph{branch instruction} if the \emph{value} register is \emph{false};
            \item perform \emph{action};
            \item \textbf{\emph{branch instruction}}: perform a \textbf{nop}
        \end{itemize}
    \end{description}

    The accompanying files to the experiment used to test this symbolic computer show some examples of programs.
    
\subsection{Initialization} \label{ComputerInitialization}
    
    For this symbolic computer to work, it needs to be trained and initialized.

    We need to:
    \begin{itemize}
        \item     Train the symbols.
        The requirements:
            \begin{itemize}
                \item a symbol for each digit, a symbol for a character that is not a digit, 
                \item a symbol for each pair of digits,
                \item a symbol for \emph{true} and a symbol for \emph{false},
                \item a symbol for special tables that we need for our example programs \emph{is-digit}, \emph{add-mod-10}, \emph{add-carry},
                \item symbols representing free memory that the program will be able to allocate and deallocate,
                \item symbols representing each code line of the hardcoded programs since we use a single pool of symbols for the code and the other symbols.
            \end{itemize}
        Thus, we must train the feedback spiking neural network representing a register on random noise to produce the prime attractors representing the symbols needed.    
        All the self-connections of the registers of the register switch box and the assignation connections involving these registers can use the same weights in an artificial setting.
        This shortcut is unrealistic for the brain but produces a similar result.
        It is much easier to do in an artificial setting than training all the registers and the assignation connection on random noise for each register.

        \item Train the \emph{opcode}, \emph{opcode-eq} and \emph{opcode-neq} clusters for the opcode symbols using sparse random noise.
        \item Train the connection between the \emph{code-c} cluster to the \emph{arg} and \emph{code2} registers and the \emph{opcode-eq} and \emph{opcode-neq} clusters according to the programs we want to encode.
        \item Create the microinstruction sequences and their effect.
        \item Bind each opcode of \emph{opcode} register to the first microinstruction of its microinstruction sequence. 
        \item Train the default value of the \emph{false} symbol to the connection from the \emph{hash} cluster to the \emph{value} register.
        \item Create the linked list of free prime attractors by initially setting the \emph{alloc} register to the \emph{false} symbol.
                Then by using the deallocation instructions to fill the \emph{alloc} linked-list with all the allocable symbols.
        \item Initialize the hash table with predefined associations used by the example programs, that is: 
            \begin{itemize}
                \item when the table and the key are symbols representing digits, bind the value to the corresponding pair of digits;
                \item when the table is the \emph{is-digit} symbol and the key is a digit symbol, bind the value to the \emph{true} symbol;
                \item when the table is the \emph{add-mod-10} symbol and the key is a pair of digits symbol, bind the value to the digit equal to the sum modulo ten of the two digits of the pair;
                \item when the table is the \emph{add-carry} symbol and the key is a pair of digits symbol, bind the value to the \emph{false} or \emph{true} symbol depending if the sum of the two digits of the pair is less than ten or not. 
            \end{itemize}
    \end{itemize}

    The symbolic machine is ready to be tested.

\subsection{Results} \label{ComputerTests}

    We tested in an experiment the implementation of this symbolic computer using spiking neurons with:
    \begin{itemize}
        \item three versions of a program that does an echo, 
        \item a program that adds two numbers represented as a sequence of digits, 
        \item and a program that counts how many there is of each digit in its input.
    \end{itemize}

    \begin{center}
        \begin{tabular}{ |c|c|c| }
        \hline
        program & result & cycles \\ 
         \hline
         echo "1234" & "1234" & 2012 \\ 
         echo2 "12345" & "54321" & 10956 \\ 
         echo3 "2345678" & "2345678" & 30993 \\ 
         add "0+1" & "1" & 17865 \\
         add "1969+1973" & "3942" & 69099 \\
         add "99995+5" & "100000" & 70219 \\
         add "21341000009+5" & "21341000014" & 65333 \\
         count-digits "0" & ":0:1:" & 28789 \\
         count-digits "212" & ":1:1::2:2:" & 51751 \\
         count-digits "2214523678703" & ":0:1::1:1::2:3::3:2::4:1::5:1::6:1::7:2::8:1:" & 169341 \\
         \hline
        \end{tabular}
    \end{center}

    Assuming that a cycle takes one millisecond, these numbers are very slow compared to human performances.
    However, there is ample room for optimizations, in which case the numbers could turn out to be very close to human performances.
  
\section{Discussion} \label{discussion}

\subsection{Computational brain model} \label{ComputationalModelOfTheBrain}
    It is interesting to look at our symbolic computer from the brain's perspective.
    The average number of outgoing connections from a neuron of $10000$ in the brain limits the number of prime attractors per register and the number of possible bindings.
    There is plenty of space for optimizations, 
    but the brain's constraint of 1 millisecond per cycle makes it very slow for sequential operations.
    Thus, the brain should rely on its pattern matching capability to compensate.
    We suggest coupling a weak symbolic computer with a powerful deep learning engine as a computational model for the brain.
    We refer to this as a neuro-symbolic computer.
    The symbolic side is limited, but maybe it is all we need to unlock the potential of the neural side.
    The symbolic side controls the execution flow, allows structure, and enforces coherence while providing molds to train the deep learning side. 
    The deep learning side accelerates the symbolic side by remembering the result of often-done operations,
    provides intuition, does recognition and performs searches.

    This separation of tasks parallels with System 1 and System 2 dichotomy described by Kahneman \cite{Kahneman11}.
    The question is whether System 2 cognition is a manifestation of this computational brain model's symbolic side.

    This work proposes multiple concepts needed to build a symbolic computer with spiking neurons and raises questions about their applicability to the brain.
    We ask:
    \begin{itemize}
        \item Are prime attractors a reality in the brain? Do the various parts of the brain use them for communication?
        \item Is the one-shot symbolic memory a reality in the brain? If so, how is it controlled? And what about unbinding?
        \item Is the control of populations of neurons for their behavior regarding Hebbian binding and unbinding a reality?
        \item Is there a linked list of unallocated prime attractors? Or is there another mechanism to allocate an attractor in the brain?
        \item Is the creation and the bookkeeping of prime attractors completed during sleep?
        \item Is there the equivalent of the register switch box in the brain? If so, how is it created?
        \item And, is the second-order network hash table a reality?
    \end{itemize}

    Another question it raises it whether these concepts could be integrated directly in the design of conventional neural networks to augment their power.

\subsection{Limits of backpropagation} \label{LimitsOfBackpropagation}

    The implementation of the symbolic computer sheds light on the limits of backpropagation to train a neural network.
    It seems unlikely that backpropagation could derive a symbolic computer like here. 
    The reliance on random noise for the prime attractors, the various components: the registers, the register switch box, the one-shot symbolic memory, the hash table, all need to be engineered and cannot be the by-product of correlating the inputs to the outputs.
    Nature chef engineer, natural evolution, needs to be involved in making these components if they are a reality for the brain.

\subsection{Improving deep learning networks through architectural priors} \label{DeepLearning2.0}

    One crucial question is how the technics developed for implementing our symbolic computer could improve the performance of deep learning networks?
    The architecture of the register switch box in section \ref{RegisterSwitchbox} uses a \emph{control} cluster to reconfigure the networks to transfer the value of one register to another one.
    We propose to use this mechanism with deep learning networks in general, noting that spiking neurons may not be needed to implement these.
    The mechanism requires assignation connections, inhibition connections, and symbol clusters that form conditional pathways carrying symbols.
    Section \ref{RegisterSwitchbox} argues that they are physiologically plausible.
    To see how these conditional pathways carrying symbols add to the design of neural networks, imagine a neural network composed of subnetworks, each having symbols as input and output. 
    Further, assume the subnetworks are linked together with these conditional pathways carrying symbols with another subnetwork controlling the open pathways.
    Training such a network implies training the subnetworks and the subnetwork controlling the pathways.
    The pathways themselves should not have their weights updated by the training as a previous independent process trained them. 
    However, they can be used to backpropagate the gradient into the other subnetworks and the subnetwork controlling the open pathways.
    The presumed advantage of such an architecture is that when training for a new problem, the network could try to find a configuration of the subnetworks already trained that solves the problem.
    If such a reconfiguration solves the problem, the training will succeed more efficiently because the network needs to find which pathways need to be open using a few critical examples instead of training a subnetwork from scratch to solve the problem using many examples.
    In that case, the network would have extrapolated arguably very well, reused the already trained subnetworks compositionally, and would have been very sample efficient.
    This benefit is at the cost of a significant architectural prior, that is, the interconnection of the subnetworks with our conditional pathways and having the subnetworks use symbols as input and output.

    More reseach is needed to explore these ideas.

\subsection{Oracle machine} \label{HybridSystems}
    Hybrid models for AI are justified if one believes in the computational model of the brain here
    due to the relative inefficiency of the implementation of symbols and their operations in neural networks 
    compared to their efficient implementation in conventional computers.

    We suggest a scheme using an oracle machine where the oracle is a deep learning engine. 
    The program queries the oracle and provides a context for an answer. 
    However, the program does not blindly trust the oracle, and the answer must be validated and tested.  
    The program informs the oracle of the result and, if required, queries it again.  
    In this way, the program trains the oracle, a neural network that initially would not be better than a random choice but would hopefully improve.      
    Take, for example, the image of an elephant.  
    The program asks the neural network to identify the image.  
    After the network informs the program that the image is an elephant, the program uses its function to validate if the image is indeed an elephant by asking the neural network questions, such as: does it have the head of an elephant? The trunk of an elephant? And so on. 
    Of course, the program should learn this function to validate if the image is an elephant.
    This scheme incidentally addresses adversarial examples.
    If the neural network wrongly identifies an elephant, then the validation procedure gives safeguards from wrongly identifying an elephant.
    If the validation procedure fails, either the neural network or the validation procedure needs an update, so the system should enter a program trying to resolve the discrepancy.

    The program could call the oracle recursively: the program could ask the oracle for a function to use to perform a task, which function, in turn, could query the oracle as well.
    
    This oracle system is different from an end-to-end learning system. 
    It is essentially a compromise: with end-to-end learning, one gains greater learnability; with a hybrid system like the oracle scheme, one gains expressivity.
    Programming is far more expressive than designing a network architecture for integrating priors. 

    Now we ask: what is the primary program for human intelligence?

\section{Conclusion}
    In this work, we introduced the concept of prime attractors held by networks called registers.
    They are instrumental in building a one-shot symbolic memory, a hash table and a register switch box.
    With these components, we built a symbolic computer out of spiking neurons and argue for its implications.
    This work shows the importance of random patterns in neural networks and their relationship to symbols.
    It also highlights the role of the number of outgoing connections per neuron, in determining various brain limits such as the maximum number of prime attractors a register can support, their minimal coverage, and the capacity of the one-shot symbolic memory.

    \textbf{Acknowledgements} We are grateful to my brother Martin Liz\'ee for his numerous helpful suggestions in improving this manuscript and my wife Ilanah Milgram for her support and corrections.

\appendix

\section{The Capacity of the One-shot symbolic memory} \label{ShorttimeMemoryCapacity}

    We proceed with an argument on the capacity of the one-shot symbolic memory.
    The argument develops in two parts.
    In the first part, we argue on the signal-noise ratio of the memory.
    In the second part, we look at the capacity of a register to recover a memory with a given signal-noise ratio.

    First, we model random sparse patterns of active neurons with random elements.
    It enables us to talk about the expected coverage of a pattern, which is at the core of the argument.

    We proceed to evaluate the one-shot symbolic memory connection's capacity by counting the ratio of the synapses in the one-shot symbolic memory connection that are opened and linking the input prime attractor to the output prime attractor. 
    We further assume that the one-shot symbolic memory connection is a random sparse matrix; that is, the synapses between the neurons of the two registers are a random selection.

    Given two registers connected with a one-shot symbolic memory connection, an input register and an output register, let $\mathbb{N}_I$ be the set of neurons of the input register and $\mathbb{N}_O$ be the set of neurons of the output register.
    Let $\mathbb{S} \subseteq \mathbb{N}_I \times \mathbb{N}_O$ be a random element representing the sparse synapses of the one-shot symbolic memory.
    
    Given $X$ an input sparse patterns modeled by a random element and $Y$ an output sparse pattern modeled by a random element as well, we define the function $\rho$ as the expected coverage.
    \begin{equation}
        \rho(X) = E(\frac{|X|}{|\mathbb{N}_I|})
    \end{equation}
    \begin{equation}
        \rho(Y) = E(\frac{|Y|}{|\mathbb{N}_O|})
    \end{equation}

    We have the following property if $X_1$ and $X_2$ are also random element representing input or output sparse patterns:
    \begin{equation}
        \rho(X_1 \cap X_2) = \rho(X_1) \rho(X_2)
    \end{equation}
    For convinience, we define the following operator giving the coverage of the union of $y$ distinct prime sparse patterns of coverage $x$:
    \begin{equation}
        \phi_x^y = 1 - (1 - x)^y < y x
    \end{equation}
    We then have that given $r$ and the $X_i$'s, all distinct and random and of coverage $c$:
    \begin{equation}
        \rho(\bigcup_{i=1}^r{X_i}) = \phi_{c}^r
    \end{equation}

    To analyze the state of the memory, we maintain the sequence of binding and unbinding events.
    Let $\mathbb{I}$ be the random elements representing the sparse random pattern defining the input prime attractors of coverage $c_1$ and $\mathbb{O}$ be the equivalent set for the output prime attractors of coverage $c_2$.
    The sequence $H_1$, $H_2$, ..., $H_i$, ... tracks all the binding operations starting from an initial state where all the one-shot symbolic memory connection synapses are zero.
    \begin{equation}
        H_i \in \{ \text{bind}, \text{unbind}\} \times \mathbb{I} \times \mathbb{O}
    \end{equation}

    Let sequence $S_0 = \emptyset$, $S_1$, $S_2$, ..., $S_i$, ... $\subseteq \mathbb{S}$ be a sequence of random elements representing the synapses opened or closed by the binding operations.
    That is:
    \begin{equation}
        H_{i+1} = (\text{bind}, X, Y) \Rightarrow S_{i+1} = S_i \cup (X \times Y) \cap \mathbb{S}
    \end{equation}
    \begin{equation}
        H_{i+1} = (\text{unbind}, X, Y) \Rightarrow S_{i+1} = S_i \setminus (X \times Y)
    \end{equation}

    We define $\Psi_i(X, Y)$ as the expected proportion of synapses in the one-shot symbolic memory connection of opened connections between $X$ and $Y$ after binding operation $H_i$.
    That is, the number of opened synapses between $X$ and $Y$ divided by the total number of synapses in the one-shot symbolic memory connection.
    \begin{equation}
        \forall X \in \mathbb{I} \land Y \in \mathbb{O}, \Psi_i(X, Y) = E(\frac{|(X \times Y) \cap S_i|}{|\mathbb{S}|})
    \end{equation}

    Given the following operator $\oplus$ defined by the following equation:
    \begin{equation}
        X_1 \oplus X_2 = X_1 \cup X_2 \text{ given } X_1 \cap X_2 = \emptyset
    \end{equation}
    We have the following identities:
    \begin{equation}
        \Psi_i(X_1 \oplus X_2, Y) = \Psi_i(X_1, Y) + \Psi_i(X_2, Y)
    \end{equation}
    \begin{equation}
        \Psi_i(X, Y_1 \oplus Y_2) = \Psi_i(X, Y_1) + \Psi_i(X, Y_2)
    \end{equation}
    
    After a \emph{bind} operation binding $Y$ to $X$, all the synapses in the one-shot symbolic memory connection between $X$ and $Y$ are opened we have:
    \begin{equation}
        \forall X \in \mathbb{I} \land Y \in \mathbb{O}, H_i = (\text{bind}, X, Y ) \Rightarrow \Psi_i(X, Y) = \rho(X) \rho(Y)
    \end{equation}
    It follows from the fact that $X$ and $Y$ are random, and the $\mathbb{S}$ is random. 
    Actually, in general, we have:
    \begin{equation}
        \forall X \in \mathbb{I} \land Y \in \mathbb{O}, \Psi_i(X, Y) \leq \rho(X) \rho(Y)
    \end{equation}
    
    Conversely after an unbinding event we have that:
    \begin{equation}
        \forall X \in \mathbb{I} \land Y \in \mathbb{O}, H_i = (\text{unbind}, X, Y ) \Rightarrow \Psi_i(X, Y) = 0
    \end{equation}
    
    A binding does not deteriorate the binding, that is:
    \begin{equation}
        \forall X_1, X_2 \in \mathbb{I} \land Y_1, Y_2 \in \mathbb{O}, H_{i+1} = (\text{bind}, X_2, Y_2 ) \Rightarrow \Psi_{i+1}(X_1, Y_1) \geq \Psi_{i}(X_1, Y_1)
    \end{equation}
    
    An unrelated unbinding does deteriorate the binding:
    \begin{equation}
    \begin{split}
        & \forall X_1, X_2 \in \mathbb{I} \land Y_1, Y_2 \in \mathbb{O}, \\ 
        & X_1 \neq X_2 \land Y_1 \neq Y_2 \land H_{i+1} = (\text{unbind}, X_2, Y_2 ) \Rightarrow \Psi_{i+1}(X_1, Y_1) \geq \Psi_{i}(X_1, Y_1) - c_1^2 c_2^2
    \end{split}
    \end{equation}
    This is because the expected ratio of synapses simulatuously linking $X_1$ to $Y_1$ and $X_2$ to $Y_2$ is given by $\rho(X_1) \rho(X_2) \rho(Y_1) \rho(Y_2) = c_1^2 c_2^2$.
    
    An unbinding with the same output prime attractor deteriorate the memory even more:
    \begin{equation}
        \begin{split}
            & \forall X_1, X_2 \in \mathbb{I} \land Y \in \mathbb{O}, \\ 
            & X_1 \neq X_2 \land H_{i+1} = (\text{unbind}, X_2, Y ) \Rightarrow \Psi_{i+1}(X_1, Y) \geq \Psi_{i}(X_1, Y) - c_1^2 c_2
        \end{split}
    \end{equation}
    This is because expected the ratio of synapses simulatuously linking $X_1$ to $Y$ and $X_2$ to $Y$ is given by $\rho(X_1) \rho(X_2) \rho(Y) = c_1^2 c_2$.

    By induction, it can be concluded that if $Y$ was bound to $X$ and not unbound since ($H_k$ being the last binding operation) and that there were $s$ unbinding operations involving $Y$ since and $t$ unbinding operations not involving $Y$ then $\Psi_k(X, Y) \geq c_1 c_2 (1 - s c_1 - t c_1 c_2)$.
    \begin{equation}
        \begin{alignedat}{2}
            & \forall X \in \mathbb{I} \land Y \in \mathbb{O}, \\ 
            & \exists i H_i = (\text{bind}, X, Y) \\
            & \land 0 = | \{ j \in (i, k] | H_j = (\text{unbind}, X, Y) \} | \\
            & \land \exists s = | \{ j \in (i, k] | H_j = (\text{unbind}, X_2, Y) \} | \\
            & \land \exists t = | \{ j \in (i, k] | H_j = (\text{unding}, X_2, Y_2) \land Y_2 \neq Y \}| \\
            & & \Rightarrow \Psi_k(X, Y) \geq c_1 c_2 (1 - s c_1 - t c_1 c_2) 
        \end{alignedat} \label{EqForgetting}
    \end{equation}

    Let $\mathbb{B}_k$ be the active bindings after $k$ binding operations:
    \begin{equation}
        \mathbb{B}_k = \{ (X, Y) \in \mathbb{I} \times \mathbb{O} | \exists i, H_i = (\text{bind}, X, Y) \land \forall j \in (i, k], H_j \neq (\text{unbind}, X, Y) \}
    \end{equation}
    There is the following identities.
    \begin{equation}
        S_k \subseteq \bigcup_{(X, Y) \in \mathbb{B}_k} (X \times Y) \cap \mathbb{S}
    \end{equation}
    \begin{equation}
        \forall X_1 \in \mathbb{I} \land Y_1 \in \mathbb{O}, \Psi_k(X_1, Y_1) \leq \sum_{(X_2, Y_2) \in \mathbb{B}_k} \Psi(X_1 \cap X_2, Y_1 \cap Y_2)
    \end{equation}
    \begin{equation}
        \Rightarrow \forall X_1 \in \mathbb{I} \land Y_1 \in \mathbb{O}, \Psi_k(X_1, Y_1) \leq \sum_{(X_2, Y_2) \in \mathbb{B}_k} \rho(X_1 \cap X_2) \rho(Y_1 \cap Y_2) \label{EqPsiSum}
    \end{equation}

    Now, assume we have bound $n = |\mathbb{B}_k|$ prime attractors of the input register to prime attractors of the output register.
    Next, assume only one prime attractor $B$ on the output register binds to an arbitrary prime attractor $A$ on the input register.
    Further assume that since the binding, $B$ has been unbound $s$ time from other attractors and there has been $t$ unbinding.
    Moreover, there is another arbitrary output prime attractor $D$ distinct from $B$ that has been bound in total by $m$ distinct input prime attractors 
    $C_1$, $C_2$, ..., $C_m$ which would have to be different than $A$. 
    We compare the ability of the network to recall $B$ over $D$ by measuring the support each would get from $A$. 
    This comparison determines if the neurons of $B$ fire before the neurons of $D$, therefore triumphing in the winner-take-all mechanism.
    We present the following:
    \begin{equation}
        \rho(A) = \rho(C_j) = c_1
    \end{equation}
    \begin{equation}
        \rho(B) = \rho(D) = c_2
    \end{equation}
    Given that $A$ is bound to $B$, and that $B$ has been unbound from other attractors $s$ times since and there had been $t$ unbinding since we have that using equation \ref{EqForgetting}:
    \begin{equation}
        \Psi_k(A, B) \geq c_1 c_2 (1 - s c_1 - t c_1 c_2)
    \end{equation}
    The part of $D$ that intersects $B$ is fully supported by $A$:
    \begin{equation}
        \Psi_k(A, D \cap B) \leq \rho(A) \rho(D \cap B) \leq c_1 c_2^2
        \label{EqDwithBs}
    \end{equation}
    The part of $A$ that intersects $\bigcup_{j=1}^m{C_j}$ fully supports $D$ and its subsets:
    \begin{equation}
        \Psi_k(A \cap (\bigcup_{j=1}^m{C_j}), D \setminus B) \leq \rho(A \cap (\bigcup_{j=1}^m{C_j})) \rho(D \setminus B)
    \end{equation}
    \begin{equation}
        \Rightarrow \Psi_k(A \cap (\bigcup_{j=1}^m{C_j}), D \setminus B) \leq c_1 \phi_{c_1}^m c_2 (1 - c_2)
        \label{EqAwithCs}
    \end{equation}
    The part of $A$ disjoint of $\bigcup_{j=1}^m{C_j}$ supports the part of $D$ disjoint of $B$ by a ratio given by
    the background noise due to the $n$ bindings which is less than $n c_1 c_2$ given that each binding affects a proportion of $c_1 c_2$ synapses.
    We use equation \ref{EqPsiSum}:
    \begin{equation}
        \Psi_k(A \setminus (\bigcup_{j=1}^m{C_j}), D \setminus B) \leq \sum_{(X, Y) \in \mathbb{B}_k} \rho(X \cap (A \setminus (\bigcup_{j=1}^m{C_j}))) \rho(Y \cap (D \setminus B))
    \end{equation}
    \begin{equation}
        \Rightarrow \Psi_k(A \setminus (\bigcup_{j=1}^m{C_j}), D \setminus B) \leq \sum_{(X, Y) \in \mathbb{B}_k} 
        \begin{cases}
            0 & \text{if } X = A \Rightarrow Y = B \\
            0 & \text{if } Y = D \Rightarrow \exists j, X = C_j \\
            c_1 \rho(A \setminus (\bigcup_{j=1}^m{C_j})) c_2 \rho(D \setminus B) & \text{otherwise} \\
        \end{cases}
    \end{equation}
    \begin{equation}
        \Rightarrow \Psi_k(A \setminus (\bigcup_{j=1}^m{C_j}), D \setminus B) \leq \rho(A \setminus (\bigcup_{j=1}^m{C_j})) \rho(D \setminus B) n c_1 c_2
    \end{equation}
    \begin{equation}
        \Rightarrow \Psi_k(A \setminus (\bigcup_{j=1}^m{C_j}), D \setminus B) \leq c_1 (1 - \phi_{c_1}^m) c_2 (1 - c_2) n c_1 c_2
        \label{EqAbutCs}
    \end{equation}
    We proceed by adding equation (\ref{EqAwithCs}) to equation (\ref{EqAbutCs}):
    \begin{equation}
        \Psi_k(A, D \setminus B) \leq c_1 c_2 [(\phi_{c_1}^m + (1 - \phi_{c_1}^m) n c_1 c_2)] (1 - c_2)
        \label{EqDbutBs}
    \end{equation}
    Then by adding equation (\ref{EqDwithBs}) to equation (\ref{EqDbutBs}), we finally get:
    \begin{equation}
        \Psi_k(A, D) \leq c_1 c_2 [c_2 + (1 - c_2) (\phi_{c_1}^m  + (1 - \phi_{c_1}^m) n c_1 c_2)]
    \end{equation}
    Now we compare the extent to which $A$ supports $D$ relative to $B$. This ratio should be significantly lower than 1 for the output
    register to favor one of $B$ over $D$. For instance, a ration of one half means that the attractor for $B$ receives
    twice the support than $D$ and should fire twice as quickly.
    \begin{equation}
        \frac{\Psi_k(A, D)}{\Psi_k(A, B)} \leq \frac{c_2 + (1 - c_2) (\phi_{c_1}^m  + (1 - \phi_{c_1}^m) n c_1 c_2)}{1 - s c_1 - t c_1 c_2}
    \end{equation}
    We simplify further and we get:
    \begin{equation}
        \Rightarrow \frac{\Psi_k(A, D)}{\Psi_k(A, B)} \leq \frac{c_2 + (1 - c_2) (n c_1 c_2 + \phi_{c_1}^m (1 - n c_1 c_2))}{1 - s c_1 - t c_1 c_2}
    \end{equation}
    \begin{equation}
        \Rightarrow \frac{\Psi_k(A, D)}{\Psi_k(A, B)} \leq \frac{c_2 + (1 - c_2) (n c_1 c_2 + m c_1 (1 - n c_1 c_2))}{1 - s c_1 - t c_1 c_2}
    \end{equation}
    \begin{equation}
        \Rightarrow \frac{\Psi_k(A, D)}{\Psi_k(A, B)} \leq \frac{n c_1 c_2 + m c_1 (1 - n c_1 c_2) + c_2 (1 - (n c_1 c_2 + m c_1 (1 - n c_1 c_2)))}{1 - s c_1 - t c_1 c_2}
    \end{equation}
    \begin{equation}
        \Rightarrow \frac{\Psi_k(A, D)}{\Psi_k(A, B)} \leq \frac{n c_1 c_2 + (1 - n c_1 c_2) (m c_1 + (1 - m c_1) c_2)}{1 - s c_1 - t c_1 c_2}
    \end{equation}
    \begin{equation}
        \Rightarrow \frac{\Psi_k(A, D)}{\Psi_k(A, B)} \leq \frac{n c_1 c_2 + m c_1 + c_2}{1 - s c_1 - t c_1 c_2}
    \end{equation}
    Given that $A$ and $D$ are arbitrary, it means that in general, the signal-noise ratio $\gamma$ of a binding network is given by:
    \begin{equation}
        \gamma \leq \frac{1 - s c_1 - t c_1 c_2}{n c_1 c_2 + m c_1 + c_2}
    \end{equation}
    Where:
    \begin{itemize}
        \item the input prime attractors have coverage $c_1$,
        \item the output prime attractors have coverage $c_2$,
        \item there are $n$ bindings with at most $m$ prime attractors bound to a given prime attractor, 
        \item since an attractor has been bound, it has been unbound at most $s$ times from another attractor,
        \item there has been at most $t$ unbindings after any active binding,
        \item and a prime attractor is bound to at most one prime attractor.
    \end{itemize}
    \begin{equation}
        \Leftrightarrow n c_1 c_2 + m c_1 + c_2 \leq \frac{1 - s c_1 - t c_1 c_2}{\gamma}
    \end{equation}
    \begin{equation}
        \Leftrightarrow n \leq \frac{\frac{1 - s c_1 - t c_1 c_2}{\gamma} - m c_1 - c_2}{c_1 c_2}
    \end{equation}
    \begin{equation}
        \Leftrightarrow n \leq \frac{1 - s c_1 - t c_1 c_2}{\gamma c_1 c_2} - \frac{m}{c_2} - \frac{1}{c_1}
    \end{equation}

    Now we ask a fundamental question: what is an acceptable signal-noise ratio?
    Part of the answer relies on the number of neurons of the input prime attractor that see each neuron of the output prime attractor. 
    We call this number $\theta$.
    That is, for any neuron of the output prime attractor, a random variable, say $S$, represents
    the number of its opened incoming synapses coming from the input cluster follows a binomial distribution of mean $\theta$ and standard deviation $\sqrt{\theta}$ given that the synapses from a random distribution and that $c_2$ is close to zero.
    The distribution of the random variable, say $T$, 
    that represents the opened incoming synapses for a neuron of the output cluster not part of the output prime attractor follows a binomial distribution with mean $\frac{\theta}{\gamma}$ and standard deviation $\sqrt{\frac{\theta}{\gamma}}$ given that $c_2$ is close to zero.
    Given that each opened synapse weights $a$, we want at $k$ steps the majority of the neurons part of the output prime attractor to have fired while very few neurons not part of it to have fired as well.
    That is, $P(kaS > 1)$ is close to one and $\frac{1-c_2}{c_2}P(kaT > 1)$ is less than one if we want the number of neurons to fire not part of the output prime attractor to be less than the number of neurons part of it.
    Now it is not clear how small this number should be and would require further study as it depends on the efficiency of the recurrent network to recover the prime attractor given noisy input.

    As an example assume that $c_1 = c_2 = \frac{30}{10000}$ and that the number of connections per neuron serving the connection is 3000 which means that $\theta = 9$.
    Further assume that $ka = \frac{1}{6}$ and that we take the signal-noise ratio $\gamma$ to be 5, which means that:
    \begin{equation}
        P(kaS > 1) = P(S > 6) \approx P(N(9;3) > 6) = 0.84          
    \end{equation}
    \begin{equation}
        \frac{1-c_2}{c_2}P(kaT > 1) = 299P(T > 6) \approx 299P(N(\frac{9}{5};\sqrt{\frac{9}{5}}) > 6) = 299 \times 0.0009 \approx 0.27          
    \end{equation}
    Furthermore if we assume that $m = 20$, $s = 30$ and $t = 10000$ our estimate for the capacity becomes:
    \begin{equation}
        n \leq \frac{1 - 30 \frac{30}{10000} - 10000 \frac{30}{10000} \frac{30}{10000}}{5 \frac{30}{10000} \frac{30}{10000}} - \frac{20}{\frac{30}{10000}} - \frac{1}{\frac{30}{10000}} \approx 11222
    \end{equation}
    This value fluctuates depending on the acceptable signal-noise ratio $\gamma$, which requires further studies.

\section{Spiking vs. non-spiking neural network}
    \label{spikingvsnonspiking}

    Now we propose to show the value for filtering noise of a spiking neural network in comparison to a non-spiking one by looking at a simplified example mathematically.

    For the spiking version of the feedback network, 
    we consider only one cluster connected to itself, with $W$ being the weight matrix, $P$ being the input pattern signal, and $s$ being a scale factor for the input.
    For simplicity, we won't have a leaking factor. 
    The expression $C_0^1(X_n) - S(X_n)$ represents the state of the neurons after the following of spiking if spiking has occurred.
    \begin{equation}
        X_{n+1} = C_0^1(X_n) - S(X_n) + WS(X_n) + sP \label{spikingRNEq}
    \end{equation}

    We follow with the equation for the non-spiking feedback network.
    Here, $W$ is the weight matrix and $P$ is the input pattern signal, while $s$ is a scale factor for the input.
    \begin{equation}
    X_{n+1} = C_0^1( WX_n + sP) \label{basicRNEq}
    \end{equation}

    We compare the behavior of the non-spiking feedback network (\refeq{basicRNEq}) to the spiking feedback network (\refeq{spikingRNEq}) one a noisy signal $P^*$
    thereby illustrating the filtering capacity of the feedback spiking network and the unsuitability of the non-spiking feedback network.
    We simplify the problem, without affecting our principal argument, by assuming that all the $A_i$'s are disjoint.
    
    A noisy signal, $P^*$, is manifactured as follow.
    \begin{equation}
        P^* = \sum_{i = 1}^l{a_i A_i} \text{ with } a_i = \begin{cases}
            1, & \text{if } i = 1 \\
            \epsilon, & \text{otherwise}
        \end{cases}
    \end{equation}
    Thus,
    \begin{equation}
        P^*[j] = \begin{cases}
            1, & \text{if } A_1[j] = 1 \\
            \epsilon, & \text{if } A_i[j] = 1 \text{ for some $i > 1$} \\
            0, & \text{otherwise}
        \end{cases}
    \end{equation}
    
    Imagine, for the example, that we train $W^*$ in the same manner as $W$ without the ReLU, and use it instead of $W$ to facilitate the computations: 
    \begin{equation}
        A_i = \frac{W^*A_i + \alpha \overrightarrow{1}}{1 + \alpha}
    \end{equation}
    \begin{equation}
        \Leftrightarrow W^*A_i = (1 + \alpha) A_i - \alpha \overrightarrow{1}
    \end{equation}

    Now use equation (\refeq{basicRNEq}) starting with $X_0 = \overrightarrow{0}$, so we have, assuming $sP^*$ is already between $\overrightarrow{0}$ and $\overrightarrow{1}$:
    \begin{equation}
        X_1 = C_0^1( W^*X_0 + sP^*) = C_0^1(sP^*) = sP^*
    \end{equation}
    \begin{equation}
        X_2 = C_0^1( W^*(sP^*) + sP^*) = C_0^1( sW^*(\sum_{i = 1}^l{a_i A_i}) + s(\sum_{i = 1}^l{a_i A_i}))
    \end{equation}
    \begin{equation}
        \Leftrightarrow X_2 = C_0^1( s(\sum_{i = 1}^l{a_i W^*A_i}) + s(\sum_{i = 1}^l{a_i A_i}))
    \end{equation}
    \begin{equation}
        \Leftrightarrow X_2 = C_0^1( s(\sum_{i = 1}^l{a_i ((1 + \alpha) A_i - \alpha \overrightarrow{1})}) + s(\sum_{i = 1}^l{a_i A_i}))
    \end{equation}
    \begin{equation}
        \Leftrightarrow X_2 = C_0^1( s(\sum_{i = 1}^l{a_i ((2 + \alpha) A_i - \alpha \overrightarrow{1})}))
    \end{equation}
    \begin{equation}
        \Leftrightarrow X_2 = C_0^1( s(((2 + \alpha) A_1 - \alpha \overrightarrow{1}) + \sum_{i = 2}^l{\epsilon ((2 + \alpha) A_i - \alpha \overrightarrow{1})}))
    \end{equation}
    Thus the value of $X_2$ is:
    \begin{equation}
        X_2[j] = \begin{cases}
            C_0^1( s(2 - (l - 1) \epsilon \alpha)), & \text{if } A_1[j] = 1 \\
            C_0^1( s(-\alpha + 2 \epsilon - (l - 2) \epsilon \alpha)), & \text{if } A_i[j] = 1 \text{ for some $i > 1$} \\
            0, & \text{otherwise}
        \end{cases}
    \end{equation}

    With many reasonable values for instance $l=100$, $\epsilon=0.3$, $c=0.01$, $\alpha=0.1$, all the components of $X_2$ are send to zero,
    that is, the inhibition response is overwhelming, thereby rendering the non-spiking feedback network useless to filter the noise.
    Note that using $W$ trained with a ReLU would not improve things; indeed, it actually augments the inhibition response.

    The spiking version of the feedback network using equation (\refeq{spikingRNEq}), as we will see, does a much better job.
    Here, only the inhibition response of the first firing neurons, those of $A_1$ here, will be acknowledged.
    The inhibition response will inhibit the competition before it can have an effect.
    The excitatory response will fire $A_1$ successively.
    Thereby continually inhibiting the part of the signal supporting the competition assuming the inhibitory response of $\alpha$ is strong enough to cover it, 
    following a winner-take-all mechanism naturally implemented by spiking neurons in this case. 
    Using equation (\refeq{spikingRNEq}) starting with $X_0 = \overrightarrow{0}$, as long as $S(X_n) = \overrightarrow{0}$, also using $W^*$ instead of $W$, we have:
    \begin{equation}
        X_{n+1} = C_0^1(X_n) - S(X_n) + W^*S(X_n) + sP^* = X_n + sP^*
    \end{equation}
    \begin{equation}
        \Rightarrow X_n = nsP^*
    \end{equation}
    \begin{equation}
        \Rightarrow X_n[j] = \begin{cases}
            ns, & \text{if } A_1[j] = 1 \\
            ns \epsilon, & \text{if } A_i[j] = 1 \text{ for some $i > 1$} \\
            0, & \text{otherwise}
        \end{cases}
    \end{equation}
    Now, as soon as $ns \geq 1$, that is: 
    \begin{equation}
        n = \lceil \frac{1}{s} \rceil
    \end{equation}
    We have that $S(X_{\lceil \frac{1}{s} \rceil}) = A_1$ assuming that $ns \epsilon < 1 $, that is:
    \begin{equation}
        \lceil \frac{1}{s} \rceil s \epsilon < 1
    \end{equation}
    And as long as we'll have that $S(X_n) = A_1$ for $n \geq \lceil \frac{1}{s} \rceil$.
    \begin{equation}
        \Rightarrow X_{n+1} = C_0^1(X_n) - S(X_n) + W^*S(X_n) + sP^* = (X_n - A_1) + W^*A_1 + sP^*
    \end{equation}
    \begin{equation}
        \Rightarrow X_{n+1} = C_0^1(X_n) - A_1 + (1 + \alpha) A_1 - \alpha \overrightarrow{1} + sP^* = C_0^1(X_n) + \alpha (A_1 - \overrightarrow{1}) + sP^*
    \end{equation}
    \begin{equation}
        \Rightarrow X_n[j] = \begin{cases}
            1 + s, & \text{if } A_1[j] = 1 \\
            ns \epsilon - (n - \lceil \frac{1}{s} \rceil) \alpha, & \text{if } A_i[j] = 1 \text{ for some $i > 1$} \\
            -(n - \lceil \frac{1}{s} \rceil) \alpha, & \text{otherwise}
        \end{cases}
    \end{equation}
    We see that if $s \epsilon < \alpha$, the components of $X_n$ not part of $A_1$ will keep diminishing and $S(X_n)$ will keep on firing only $A_1$.
  
    Of course, the situation might not always be clear-cut.
    Not all the neurons of $A_1$ might spike, 
    slowing down the repeated activation of $A_1$ and enabling the competition to fire as well.
    In this case, the potential attractors compete to be the emergent one. 
    There is also the possibility that if too many prime attractors fire simultaneously, 
    we end up in the situation of the non-spiking feedback network. 
    Nevertheless, in practice, if the signal supports $A_1$ with a significantly higher rate than the competition, it emerges from the networks with the noise considerably reduced.

    \pagebreak

    \printbibliography 

    \listoffigures

 \end{document}